\pdfoutput=1

\documentclass[11pt]{article}

\usepackage[]{ACL2023}

\usepackage{times}
\usepackage{latexsym}

\usepackage[T1]{fontenc}

\usepackage[utf8]{inputenc}

\usepackage{microtype}

\usepackage{inconsolata}

\usepackage{graphicx}
\usepackage{subcaption}
\usepackage{multirow}
\usepackage{booktabs}
\usepackage{makecell}
\usepackage{tabularx}
\usepackage{accents}
\usepackage{amssymb}
\usepackage{pifont}

\usepackage{longtable}
\usepackage{tablefootnote}

\newcommand{\eg}{e.g.,}
\newcommand{\ie}{i.e.,}
\newcommand{\etc}{etc.}

\newcommand\figref[1]{Figure~\ref{#1}}

\newcommand\tabref[1]{Table~\ref{#1}}

\newcommand\secref[1]{\S\ref{#1}}

\newcommand\appref[1]{Appendix~\ref{#1}}

\newcommand{\greencheckmark}{{\color{green}{\checkmark}}}
\newcommand{\redxmark}{{\color{red}{\ding{55}}}}

\newcommand{\Sref}[1]{\S\ref{#1}}

\newcommand{\fakeparagraph}[1]{\vspace{1mm}\noindent\textbf{#1}}

\usepackage{collcell}
\newcommand{\mytexttt}[1]{\raggedright\texttt{#1}}
\newcolumntype{M}[1]{>{\collectcell\mytexttt}p{#1}<{\endcollectcell}}

\title{Stumbling Blocks: Stress Testing the Robustness of Machine-Generated Text Detectors Under Attacks}

\newcommand{\affilXJ}{\ensuremath{^\clubsuit}}
\newcommand{\affilUW}{\ensuremath{^\blacklozenge}}
\newcommand{\affilJHU}{\ensuremath{^\spadesuit}}
\newcommand{\affilPKU}{\ensuremath{^\blacktriangle}}

\author{ Yichen Wang\affilXJ \quad 
         Shangbin Feng\affilUW \quad
         Abe Bohan Hou\affilJHU \quad 
         Xiao Pu\affilPKU \\
         \textbf{Chao Shen}\affilXJ \quad
         \textbf{Xiaoming Liu}\affilXJ \quad
         \textbf{Yulia Tsvetkov}\affilUW \quad
         \textbf{Tianxing He}\affilUW \\
         \affilUW Paul G. Allen School of Computer Science \& Engineering, University of Washington \\
         \affilXJ Xi'an Jiaotong University \quad
         \affilJHU Johns Hopkins University \quad 
         \affilPKU Peking University\\
         {\tt yichen.wang@stu.xjtu.edu.cn \quad goosehe@cs.washington.edu} \\
    }

\begin{document}
\maketitle
\begin{abstract}
The widespread use of large language models (LLMs) is increasing the demand for methods that detect machine-generated text to prevent misuse.
The goal of our study is to stress test the detectors' robustness to malicious attacks under realistic scenarios.
We comprehensively study the robustness of popular machine-generated text detectors under attacks from diverse categories:  \textit{editing}, \textit{paraphrasing}, \textit{prompting}, and \textit{co-generating}.
Our attacks assume limited access to the generator LLMs, and we compare the performance of detectors on different attacks under different budget levels.
Our experiments reveal that almost \emph{none} of the existing detectors remain robust under all the attacks, and all detectors exhibit different loopholes. Averaging all detectors, the performance drops by 35\% across all attacks. 
Further, we investigate the reasons behind these defects and propose initial out-of-the-box patches to improve robustness.\footnote{Preprint. We release code and data at \url{https://github.com/YichenZW/Robust-Det}. 
Yichen Wang and Tianxing He are the corresponding authors.}
\end{abstract}

\newcolumntype{L}[1]{>{\raggedright\arraybackslash}m{#1}}
\begin{table*}[h]
\centering
\renewcommand\arraystretch{1.2}
\resizebox{\linewidth}{!}{
\begin{tabular}{ l l c c c L{9cm} }
\toprule
{\textbf{\texttt{Attack} Category}} & {\textbf{Method}} & {\textbf{Model-Free?}} & {\textbf{Level}} & {\textbf{Access}} & {\textbf{Detailed Descriptions}} \\
\midrule
\multirow{3}{*}[-2em]{\begin{tabular}{@{}l@{}}\textbf{Editing} \\ (\Sref{editing}) \\ \small{\texttt{post-generation}}\end{tabular}} & Typo Insertion &   \greencheckmark & Character & None & Create typos by inserting, deleting, substituting, and transposing mainly. \\

& Homoglyph Alteration&   \greencheckmark & Character & None & Change English characters into visually similar Unicodes, \eg{} Cyrillic characters. \\
& Format Character Editing &  \greencheckmark & Character & None & Change or insert formatting characters, including  zero-width whitespace \texttt{\textbackslash{}u200B} insertion, and shift character editing, \eg{} \texttt{\textbackslash{}n, \textbackslash{}r}, \texttt{\textbackslash{}u000B} (vertical tab), \etc{}\\

\midrule
\multirow{4}{*}[-2em]{\begin{tabular}{@{}l@{}}\textbf{Paraphrasing} \\ (\secref{paraphrasing}) \\ \small{\texttt{post-generation}} \end{tabular}} & Synonyms Substitution & opt \greencheckmark{} or \redxmark{} & Word & None & For model-free (\greencheckmark{}) setting, retrieve a synonym from a static dictionary; for model-based (\redxmark{}) setting, utilize a LLM to generate synonyms list given context.\\
& Span Perturbation & \redxmark{} & Span & None & Use a masked LM \citep{raffel2020exploring} to rewrite spans of tokens by masked filling. \\
& Inner-Sentence Paraphrase & \redxmark{} & Inner-Sent. & None & Use Pegasus \citep{zhang2020pegasus} to paraphrase each sentence of the text and then join them. \\
& Inter-Sentence Paraphrase & \redxmark{} & Inter-Sent. & None & Paraphrase with Dipper \citep{krishna2023paraphrasing}, a paragraph-level paraphraser that can re-order, split, and merge sentences meanwhile paraphrasing each sentence.  \\

\midrule
\multirow{3}{*}[-2em]{\begin{tabular}{@{}l@{}}\textbf{Prompting} \\ (\secref{prompting}) \\ \small{\texttt{pre-generation}} \end{tabular}} 
& Prompt Paraphrasing & \redxmark{} & Inter-Sent. & Prompting & Paraphrase the raw prompt before generation using Pegasus. \\
& In-Context Learning & \redxmark{} & Inter-Sent. & Prompting & Given the example of HWT and MGT as positive and negative demonstrations when generating MGT on the same prompt.\\
& Character-Substituted Generation & \redxmark{} & Inter-Sent. & Prompting & Prompt to ask the model to generate the text with specific character substitution criteria and recover the output after finishing the whole generation. \\

\midrule

\multirow{2}{*}[-1em]{\begin{tabular}{@{}l@{}}\textbf{Co-Generating} \\ (\secref{co-gen}) \\ \small{\texttt{on-generation}} \end{tabular}} 
& Emoji Co-Generation & \greencheckmark{} & Inter-Sent. & Decoding & Compulsorily generate or insert an emoji after finishing each sentence while recurrent generation and remove all the emojis after finishing the whole text.\\
& Typo Co-Generation & \greencheckmark{} & Inter-Sent. & Decoding & Preset character substitution rules and execute the rules when finishing sampling each token and recover them after finishing the whole text generation.\\

\bottomrule

\end{tabular}
}
\caption{\textbf{Overview of the attacks.} `Model-Free' means whether the attacker is free from using any additional language model or not. `Access' indicates the access to the generator needed when doing the attack (details in \secref{attacks} and examples in \tabref{tab:attack_exmaple}).} 
\vspace{-15pt}
\label{tab:attacks}
\end{table*}

\section{Introduction}

LLMs are becoming increasingly adopted in information seeking scenarios, assistive writing, translation, mental health support, and many more \citep{zhao2023survey}.
Their evolving capabilities to generate human-like and persuasive language raise wide concerns about misuse, \eg{} deception, academic misconduct, and disinformation \citep{zellers2019defending,weidinger2021ethical, kumar2022language, feng2024does}, and it becomes harder for humans to distinguish machine-generated texts (MGT) from human-written texts (HWT) \citep{dugan2023real}. 
As a result, much recent work focus on automatic MGT detection to mitigate the risks \citep{liu2022coco, mitchell2023detectgpt, kirchenbauer2023watermark, mao2024raidar}.


In this work, we focus on potential malicious attacks that attempt to deceive the detector using various attack strategies.
Existing works on this topic mostly focus on the robustness of specific detectors or particular attack methods.
For example, \citet{zhang2023assaying} specifically evaluate the topic-shifting attack for metric-based detectors, \citet{liu2022coco} assess the token editing attack for model-based detectors, and \citet{krishna2023paraphrasing} particularly study the paraphrasing-based attack, \etc{}
To the best of our knowledge, in the literature, there is no thorough comparative evaluation of robustness of machine-generated text detectors against malicious attacks, covering a wide range of detectors and attacks. 


\begin{figure}[t]
  \centering
  \includegraphics[width=1.0\linewidth]{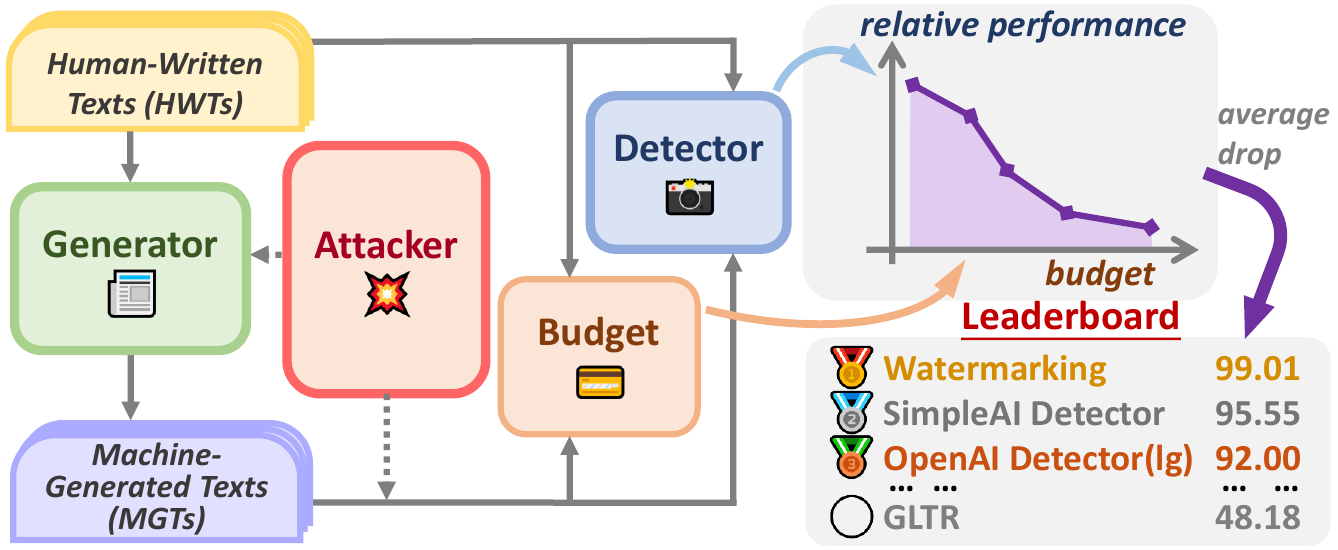}
  \caption{\textbf{Pipeline of the study.} The attacks are carried out on the machine-generated texts before, during, or after generation. Each attack is applied with different perturbation levels, denoted as budgets (\secref{budget}).}
  \label{fig:intro} 
\end{figure}

With this goal, we study the robustness of 8 prevalent MGT detectors from 3 categories under 12 realistic attacks (\secref{attacks}, \tabref{tab:attacks}), including editing, paraphrasing, prompting, co-generating, \etc{} The majority of the attacks in this paper are proposed or attempted for the first time.
The implications of our study are: 
\textit{a}) For users, our results identify and compare the limitations of each detector, which could help make better-informed choices of detectors based on individual scenarios and requirements. 
\textit{b}) For developers, our implemented suite of attacks could help them comprehensively and comparatively evaluate the robustness of detectors, and importantly, identify potential weaknesses.

We design a suite of attacks (\secref{attacks}) on MGT detectors under realistic scenarios with a range of methodologies (\tabref{tab:attacks}).
For a fair comparison across detectors and attacks, we utilize a series of metrics to measure the perturbation level of each attack, which we term ``budget'' (\secref{budget}). 
Strikingly, our experiments (\secref{overall_message}) reveal that \textbf{almost none of the existing detectors remains robust under all the attacks}, showing a variety of potential weaknesses or loopholes. 
For example, about 2 to 6-character editing by typo insertion can severely deceive metric-based detectors, such as DetectGPT \citep{mitchell2023detectgpt}, to perform worse than a random prediction (\secref{editing}), \etc{}
Hence, we view the attacks as the \textit{stumbling blocks} for current MGT detectors toward robustness. Moreover, we interpret the reasons behind the detectors' weaknesses under attacks, and we further introduce out-of-the-box patches with inferior performance in some scenarios (further defense discussed in \appref{app:future_work_d}).

We build a robustness leaderboard (\tabref{tab:leaderboard}, and the pipeline is illustrated in \figref{fig:intro}) by averaging results from different attacks. We find that \textbf{watermarking} \citep{kirchenbauer2023watermark} \textbf{performs best for robust MGT detection to its applicable attacks.}\footnote{Watermarking requires logit-level access to the generator model and has the risk of negatively impacting text quality.}
Next, \textbf{model-based detectors are more robust than metric-based ones in most cases.}
Overall, this study aims to raise awareness of the detection vulnerabilities and the urgency of more robust methodologies, thereby turning the \textit{stumbling blocks} into \textit{stepping stones}.


\vspace{-6pt}
\section{Problem Formulation}
\vspace{-7pt}
\fakeparagraph{Threat Model.\label{sec:for-overall}}
\figref{fig:intro} shows the overall pipeline. There are three roles in the problem: \textit{generator} (\secref{generators}), \textit{detector} (\secref{detectors}), and \textit{attacker} (\tabref{tab:attacks}, \secref{attacks}). 
The task for the detector is to classify whether a given piece of text is human-written (HWT) or machine-generated (MGT) from the generator LM. 
In the attacked scenario, before the MGT is sent to the detector, an attacker could tamper with the text or the generator, attempting to deceive the detector into classifying the MGT as HWT.
We compute the \textit{budget} (\secref{budget}) of each attack to measure its impact on text quality and semantics.

\vspace{-1pt}
\fakeparagraph{Scope.\label{sec:scope}}
For a realistic scenario, we set the scope of our robustness evaluation under attack as follows:
    \textit{\textbf{i}}) We assume that the attacker does \textbf{not} have any knowledge or access to the \textbf{detectors}. 
    \textit{\textbf{ii}}) The attacker only has \textbf{limited} access to the \textbf{generators}: 
    We assume to have prompting access with tunable sampling hyper-parameters for the following reason: currently, most top-performing LLMs accessible to users are closed-source (\eg{} GPT-4, Claude), to which we only have API access or a panel including a prompt input and sampling settings \citep{OpenAI2022ChatGPT}. Due to the same reason, adversarial attacks \citep{li2018textbugger, le-etal-2022-perturbations} are not covered in this study.
    \textit{\textbf{iii}}) For a holistic comparison, we apply each attack on different perturbation levels (e.g., number of typos), termed as budgets (\secref{budget}). 
\vspace{-6pt}

\section{Generator and Detector}

\subsection{Generators\label{generators}}

We select GPT-2 XL (1.5B) \citep{Radford2019LanguageMA}, GPT-J (6B) \citep{gpt-j}, and LlaMA-2 (7B-hf) \citep{touvron2023llama2} as the representative open-source generators, and Text-Davinci-003 \citep{OpenAI2022Text-Davinci-003} and GPT-4 \citep{openai2023gpt4} as the closed-source generator representatives. 
\textbf{All the generators shared similar results under attacks} (\appref{app:acs_gens}).
We select GPT-J (6B) as the default generator to show the results in \secref{attacks} if unspecified (we empirically find stronger generative LMs are not as well good at detecting when used for metric-based detectors).
The results of LlaMA-2 and GPT-4 will be additionally shown in \appref{app:acs_gens} and \secref{attacks}. 
For closed-source generators, some of the detectors can not be applied due to the requirement of white-box parameters.

\subsection{Detectors\label{detectors}}

Current MGT detectors could be classified into 3 high-level categories, as we introduce below. We include representative detectors from each category for our evaluation. 

\fakeparagraph{Metric-Based Detector} relies on the inferenced log-probability from the generator LLM, and adopts a threshold for classification.\footnote{The setting of threshold largely impacts the detection accuracy, but it is out-of-the-scope of this paper's focus. Thus, we mainly use threshold-free metrics (\eg{} \textit{AUC ROC} and \textit{TPR@FPR}) in experiments (detailed in \secref{metrics}).} Detectors for this type does not require any training. 

\textit{GLTR} \citep{gehrmann2019gltr, solaiman2019release} using the average of the next-token probability to determine whether an input text is MGT. Texts with high average probability are classified as MGTs.

\textit{Rank} and \textit{LogRank} \citep{solaiman2019release, mitchell2023detectgpt} using the averaged rank and log-rank of next-token probability for detection respectively.


\textit{DetectGPT} \citep{mitchell2023detectgpt} stands as the pioneering work of using perturbation as a comparison to original texts to enhance metric-based detection. Perturbation here refers to rewriting or substituting spans of tokens using a mask-filling LM (\ie{} T5-small \citep{raffel2020exploring}). It poses that perturbed MGTs tend to have lower log probabilities compared to the original samples under the base LM, while perturbed HWTs may be at about a similar level to the origin.
\citet{bao2023fast, su2023detectllm, Liu2024DoesF, mao2024raidar} further follow up DetectGPT. 

We apply the white-box setting to the metric-based detectors, where full knowledge (\eg{} which LLM generated the texts) and access ( including the parameters of the generator LLM) are given to the detectors. The reason is that those detectors require the generator LLM as the base model to compute the metrics.

\fakeparagraph{Fine-Tuned Detector} is trained on a pretrained language model (PLM) in a supervised method with a classification loss.

\textit{OpenAI Detector} \citep{solaiman2019release} is a model to detect GPT-2 generation by fine-tuning a RoBERTa \citep{liu2019roberta} model. We evaluate both the base size (125M) and the large size (355M) models.

\textit{SimpleAI Detector} \citep{guo2023close} is a detector mainly for distinguishing ChatGPT, using the HC3 QA dataset \citep{guo2023close} to fine-tune a RoBERTa model.
 
\textit{Fine-tuned DeBERTa} is the model we fine-tuned on our generation data, representing an in-domain setting. We use DeBERTa-v3-base \citep{he2021debertav3} as the base model.\footnote{We have also tried other base models, \eg{} BERT \citep{devlin2018bert}, RoBERTa, ELECTRA \citep{clark2020electra}, \etc{} The selection of base model does not impact the overall trend of the findings, and the gap on the absolute detection accuracy is within 2\%.} 

Compared with OpenAI and SimpleAI Detectors as off-the-shelf models, our fine-tuned DeBERTa is relatively in-domain since it is solely fine-tuned on the dataset from the same generator and within the same topic domain as the test set.
All the fine-tuned detectors are under the black-box setting, which means they have no knowledge or access to the generator LLM but only the generated dataset.

\fakeparagraph{Watermark-Based Detector} adds algorithmically detectable signatures into texts during generation. \citet{kirchenbauer2023watermark} is a representative approach, which adds a token-level bias in the decoding stage (represented as \textit{Watermark} afterward).
This work is followed up by \citet{zhao2023provable, christ2023undetectable, kuditipudi2023robust, hou2023semstamp}.
All watermark-based detectors are under the white box setting, where they have all the knowledge and access to the generator LLM.

We follow the recommended configurations for most detectors. Detailed hyperparameters are included in the \appref{app:detail-hyper}.

\section{Budget of Attacks\label{budget}}

As stated in \Sref{sec:scope}, to measure the perturbation level of attacks on the generated texts, we utilize a series of text generation evaluation metrics as the budget of attacks, covering syntactic- or semantic-level perturbation. A strong attack should induce large detection performance degradation with a relatively small budget.

For the editing attacks, we use \textit{Levenshtein Edit Distance} \citep{Levenshtein1965BinaryCC} as the major budget, which is the minimum number of single-character edits, including insertions, deletions, and substitutions. A larger distance represents a larger attack budget. Additionally, we also record \textit{Jaro Similarity} \citep{Jaro1989AdvancesIR}.\footnote{The edit distance, Jaco similarity, and cosine similarity are implemented based on the \texttt{string2string} \citep{suzgun2023string2string} package.}

To measure the quality of texts under the attacks that change the semantic meaning (\eg{} prompting attacks and co-generating attacks), we utilize \textit{Perplexity} under LlaMA-7B-hf \citep{touvron2023llama} and \textit{MAUVE} \citep{pillutla2021mauve}.
We use MAUVE to compare the distribution gap between MGTs and HWTs. MGTs are used to estimate the model distribution, and HWTs are used to estimate the target distribution (the setting is abbreviated as `M2H'). Lower Perplexity or higher MAUVE (M2H) represents better quality and a smaller budget. \tabref{tab:base_budget} shows the unattacked value for reference.

\begin{table}[t]
\centering
\resizebox{\linewidth}{!}{
    \begin{tabular}{cc | cc}
        \toprule
         \multicolumn{2}{c |}{\textbf{\textit{Unwatermarked}}}&  \multicolumn{2}{c}{\textbf{\textit{Watermarked}}}\\
         \midrule
         \textit{PPL}&  \textit{MAUVE(M2H)}&  \textit{PPL}& \textit{MAUVE(M2H)}\\
         1.930 $\pm$ 0.386&  0.9444&  2.119 $\pm$ 0.524  & 0.9639\\
         \bottomrule
    \end{tabular}
    }
    \caption{The unattacked value of average Perplexity and MAUVE (M2H) as the base point. Notably, for the watermark-based detector, the reference texts for budget computation are watermarked MGTs instead of the original unwatermarked MGTs.}
    \label{tab:base_budget}
\end{table}

 For the attacks that do not change semantics meaning, \eg{} paraphrasing, we use \textit{BERTScore} \citep{zhang2019bertscore} as the major metric for the budget. We utilize it to compare the similarity between MGTs after the attack to MGTs before the attack. 
 In this scenario, attacked MGTs are the candidates for BERTScore, while unattacked MGTs are the reference (the setting is abbreviated as `A2B').
 The BERTScore we used is rescaled. A larger BERTScore (A2B) value means a smaller budget in the attack.
Besides, we also record \textit{BARTScore} \citep{yuan2021bartscore} and \textit{Cosine Similarity}, which shows equivalent results.

See \tabref{tab:budgets} for more details on the metrics for the attack budget. 
\appref{app:acs_budgets} show the correlation among all metrics and they show highly similar trends of attacked performance. 




\section{Experiment Setting}

\subsection{Data Setting}

Following the setting of \citet{pu2023zero}, we generate News-style texts with a proper sampling strategy for each generator, detailed in \appref{app:dataset_gen}. Our study can be readily applied to data from other domains.
The prompts used for MGT generation are the first 20 tokens of HWTs in the dataset.
The setting of the sampling strategy aims to prevent repetition, measuring by duplicate n-grams \citep{welleck2019neural}.
The size of the training, evaluation, and testing set is 8,000, 1,000, and 1,000, respectively, with balanced labels.

\subsection{Metrics for Detector Performance\label{metrics}}

The metrics we use to evaluate detection performance are binary classification metrics \textit{AUC ROC} and \textit{TPR@FPR}. 
AUC ROC is the area under the receiver operating characteristic curve.
TPR@FPR is the true positive rate when the false positive rate is at a specific percentage. Under our setting, it is equivalent to \textit{Attack Success Rate (ASR)} \citep{tsai-etal-2019-adversarial-attack}. We mainly show \textit{TPR@FPR=5\%}, and TPR@FPR=10\% and =20\% are additionally recorded in the \appref{app:acs_metrics}. 
We do not involve \textit{Accuracy} and \textit{F1-score} because those metrics are dependent on the setting of the threshold for metric-based detectors, which could be biased in the comparison.
Notably, we report all the metrics of attacked scenarios in relative value to the unattacked performance (reported in \tabref{tab:unattacked}) for clearer comparison.

\section{Attacks and Results\label{attacks}}
In this section, we describe the attack methodologies and results divided by attack category. We view the degraded performance under attacks of various detectors as \textit{stumbling blocks} to robust MGT detection. 
Further, we analyze the defects and propose defense patches in each category to explore the potential of \textit{turning stumbling blocks into stepping stones}.
\tabref{tab:attacks} is an overview of all attacks and \tabref{tab:attack_exmaple} shows some examples.

\begin{table}[t]
\centering
\resizebox{\linewidth}{!}{
\renewcommand\arraystretch{1.1}
\begin{tabular}{l c c c c c}  
\toprule
\multicolumn{6}{c}{\cellcolor[HTML]{E5E7E9}{\textbf{Leaderboard: MGT Detector Robustness}}} \\
\midrule
\texttt{\textbf{Detector}}& \textit{Edit} & \textit{Para.} & \textit{Prompt} & \textit{CoGen.} & \textbf{\textit{Avg.}} \\
\midrule
\textcolor[HTML]{FFB300}{Watermark} & \textbf{99.86} & \textbf{97.17} & - - & \textbf{99.99} & \textbf{99.01}* \\
\textcolor[HTML]{90A4AE}{SimpleAI Det.} & \textbf{108.1} & 97.51 & 81.58 & 95.04 & \textbf{95.55} \\
\textcolor[HTML]{AF601A}{OpenAI Det.-Lg} & \cellcolor[HTML]{EB984E}{57.77}& \textbf{97.84} & \textbf{105.2} & \textbf{107.2} & 92.00 \\
\textit{Model. Avg.} & 76.65 & 92.08 & 97.57 & 92.22 & 89.63 \\
F.t. DeBERTa& 104.1 & 81.49 & 99.09 & \cellcolor[HTML]{EB984E}{64.28}& 87.24 \\
OpenAI Det.-Bs& \cellcolor[HTML]{EB984E}{36.63}& 91.46 & 104.4 & 102.4 & 83.71 \\
DetectGPT-1d & \textbf{74.82} & \textbf{75.32} & \textbf{102.8} & \cellcolor[HTML]{EB984E}{\textbf{66.46}}& \textbf{79.85} \\
DetectGPT-10d & \cellcolor[HTML]{EB984E}{62.67}& 64.40 & 97.68 & \cellcolor[HTML]{EB984E}{49.78}& \cellcolor[HTML]{EB984E}{68.63}\\
DetectGPT-10z & \cellcolor[HTML]{EB984E}{56.41}& \cellcolor[HTML]{EB984E}{59.73}& 93.88 & \cellcolor[HTML]{EB984E}{43.08}& \cellcolor[HTML]{EB984E}{63.28}\\
\textit{Metric. Avg.} & \cellcolor[HTML]{EB984E}{51.82}& \cellcolor[HTML]{EB984E}{61.89}& 91.26 & \cellcolor[HTML]{EB984E}{33.49}& \cellcolor[HTML]{EB984E}{59.62}\\
LogRank & \cellcolor[HTML]{EB984E}{41.76}& \cellcolor[HTML]{EB984E}{58.38}& 84.44 & \cellcolor[HTML]{EB984E}{11.20}& \cellcolor[HTML]{EB984E}{48.95}\\
Rank & \cellcolor[HTML]{EB984E}{36.46}& \cellcolor[HTML]{EB984E}{57.68}& 81.00 & \cellcolor[HTML]{EB984E}{20.08}& \cellcolor[HTML]{EB984E}{48.81}\\
GLTR & \cellcolor[HTML]{EB984E}{38.82}& \cellcolor[HTML]{EB984E}{55.80}& 87.79 & \cellcolor[HTML]{EB984E}{10.32}& 48.18 \\
\bottomrule

\end{tabular}
}
\caption{\textbf{The overall robustness leaderboard of MGT detectors} by averaging the relative AUC ROC percentage across all attack budget levels in \secref{attacks}, ranking downwards by the overall average. 
`\textit{Metric. Avg.}' and `\textit{Model. Avg.}' represent the average performance of metric-based and model-based detectors. Bolding indicates the best performance in each detector category, and worse performance with drops larger than 70\% are in orange.}
\label{tab:leaderboard}
\end{table}

\begin{table}[h]
\centering
\resizebox{\linewidth}{!}{
\renewcommand\arraystretch{1.1}
\begin{tabular}{l c c c c c }  
\toprule
\multicolumn{6}{c}{\cellcolor[HTML]{E5E7E9}{\textbf{Absolute MGT Detector Performance w/o Attack}}} \\
\midrule
\texttt{\textbf{Detector}} & \textit{AUC} & \textit{TF=5} & \textit{TF=10} & \textit{TF=20} & \textit{ACC} \\
\midrule
GLTR & 84.46& 39.00& 53.40& 71.60& 76.00
\\
Rank & 68.13& 22.60& 35.60& 46.80& 63.60
\\
LogRank & 87.36& 50.00& 65.60& 78.20& 79.00
\\
Entropy & 51.84& 7.60& 14.60& 26.40& 50.80
\\
\midrule
DetectGPT-1d & 68.66& 15.80& 27.40& 45.80& 62.10
\\
DetectGPT-10d & 83.12& 21.60& 43.80& 71.20& 75.80
\\
DetectGPT-10z & 85.16& 30.80& 50.80& 73.20& 76.20
\\
\midrule
OpenAI Det.-Bs & 83.12& 42.40& 56.20& 69.00& 75.00
\\
OpenAI Det.-Lg & 88.55& 53.60& 65.60& 78.00& 79.00
\\
SimpleAI Det. & 87.98& 81.20& 82.60& 84.60& 84.40
\\
F.t. DeBERTa & 91.90& 5.40& 49.20& 99.60& 88.80
\\
\midrule
Watermark & 99.94 & 99.80 & 99.80 & 99.80 & 99.99\\
\bottomrule
\end{tabular}
}
\caption{\textbf{The performance of the detectors in the unattacked scenario (absolute value).} For short, `AUC' is ROC AUC, `TF=5' is TPR@FPR=5\%, `ACC' is Accuracy, `Det.' is Detector, and `F.t.' is Fine-tuned.}
\label{tab:unattacked}
\end{table}

\begin{figure*}[t]
\centering
\includegraphics[width=\textwidth]{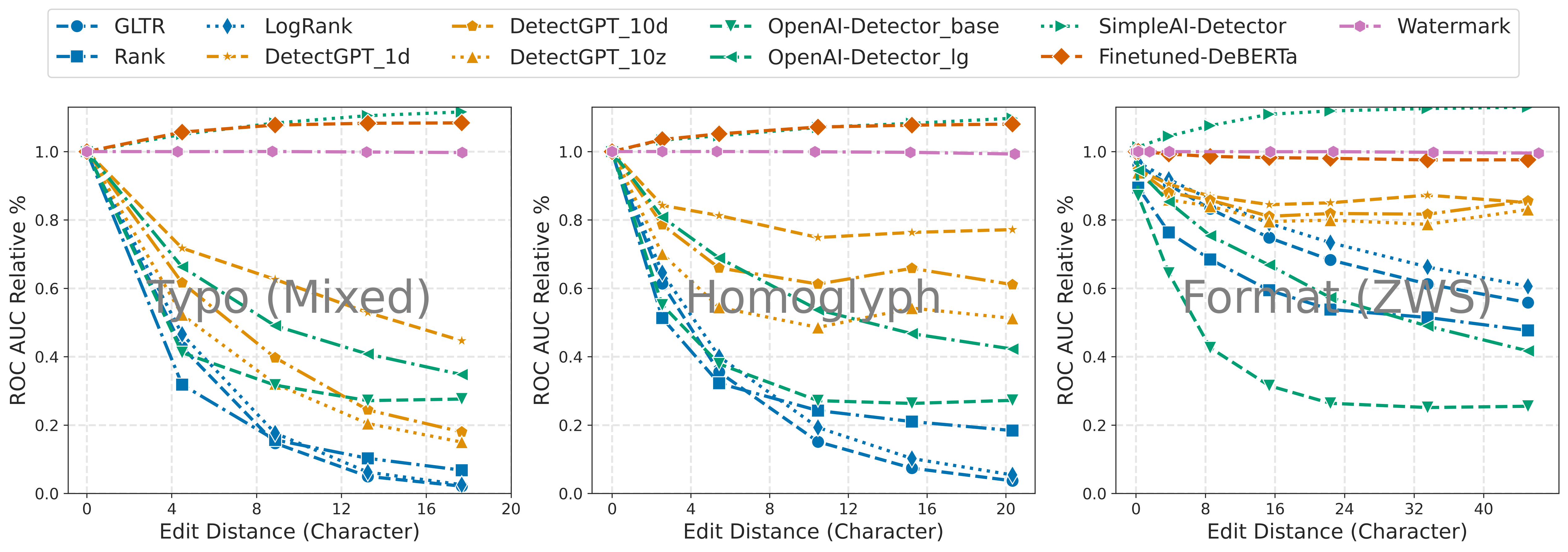}
\caption{\textbf{Performance drop of the detectors under the editing attacks.} We show the mixed setting for typo insertion, and the zero-width whitespace setting (ZWS for short) for format character editing. The budget on the x-axis is the edit distance at character level ($\uparrow$ a larger number represents a stronger attack). The color of dashed lines indicates the category of detectors.\footnotemark}
\label{fig:editting}
\end{figure*}
\footnotetext{The x-ticks in format (ZWS) character editing is twice the ones in typo and homoglyph because the Unicode is 2 bytes when computing edit distance.}

\subsection{Overall Message\label{overall_message}}

For readers who want a high-level overview of our findings, we show the overall results and messages ahead here by aggregating results from all types of attacks covered in our work.
We will introduce and discuss the detailed attacks and results in the following subsections (\secref{editing} - \secref{co-gen}).

\fakeparagraph{Leaderboard.}
Overall, we build a leaderboard of detector robustness averaging all the performance datapoints under attacks. The relative AUC ROC under attack\footnote{`Relative AUC ROC under attack' is the percentage of the AUC ROC in attacked scenarios divided by the unattacked AUC ROC, to show the relative performance drop of the detectors under attack. Detailed in \secref{metrics}.} are as shown in \tabref{tab:leaderboard}. A high relative AUC ROC means that the detector is robust to the attack.
According to the leaderboard, \textbf{watermarking is most robust to accessible attacks}\footnote{Some prompting attacks can not be done on watermark since in need of white-box models and compatibility to the watermarking decoding.}.
Following, SimpleAI Detector and OpenAI Detector (large) rank second and third. 
Moreover, \textbf{model-based detectors are more robust than metric-based detectors} in most cases. 
Additionally, we report the absolute performance of the detectors without attacks in \tabref{tab:unattacked}, which should also be considered while selecting suitable detectors.

\fakeparagraph{Detector Defect Review.}
We summarize the defect for each detector as follows. 
GLTR, Rank, and LogRank have an average performance drop of 51.35\% under all attacks, especially not robust to editing, paraphrasing, and co-generating attacks.
In comparison, DetectGPT shows better robustness (average 29.41\% drop), especially on paraphrasing and co-generating attacks. 
Among fine-tuned detectors, the strongest attack method varies.
SimpleAI Detector drops performance on paraphrasing and prompting attacks, OpenAI Detectors drop on editing, and F.t. DeBERTa performs worse on co-generating while it keeps decent robust to other attacks. We empirically find a larger model size of OpenAI Detectors eases the robustness drawback.
Notably, watermarking is robust to all applicable attacks, but it requires decoding-time access to the generator compared with other detectors. 

\subsection{Editing Attacks\label{editing}}

The first attack type we explore is the editing attacks, which are applied to the generated texts by minor editing at the character level without any change in semantics at the post-generation stage. 
Thus, editing attacks are at a low granularity. Some of the attacks might cause the text to lose minor quality and readability. Below, we will introduce three attack types.

\subsubsection{Approaches}

\fakeparagraph{Typo Insertion\label{typo}} 
intentionally adds a few typos into generated texts.
We consider four main kinds of typos in English keystroke scenarios: insertion, deletion, substitution, and transposition \citep{Kukich1992TechniquesFA}. 
Aside from testing on each kind, we propose a mixed typo insertion to mimic the realistic scenario according to the distribution investigated by \citet{Baba2012HowAS}.\footnote{substitution 55.6\%, insertion 20.3\%, transposition 1.1\%, deletion 23.0\%.} 
Also, we additionally take letter frequency into account when selecting the characters to be attacked \citep{Pavel2000letterfreq}.

\fakeparagraph{Homoglyph Alteration\label{homoglyph}}
uses graphemes, characters, or glyphs with visually identical or very similar shapes but different meanings for imperceptible replacements. It is first introduced in the cyber security domain \citep{10.1145/503124.503156}.
We use VIPER \citep{eger2019text} (Visual Perturb) Easy Character Embedding Space (ECES) to get the best homoglyph alternative of the selected character.

\fakeparagraph{Format Character Editing\label{format}}, also named Discreet Alteration \citep{kirchenbauer2023watermark}, uses special escape characters and format-control Unicodes as human-invisible disruptions to deceive detectors. See details in \appref{app:homoglyph}.

We do all the editing on the character level, and the budget is measured by edit distance. 
Also, we do at most one edit per word.

\subsubsection{Results and Analysis}
As shown in \figref{fig:editting}, all metric-based detectors and some fine-tuned ones drop dramatically, while only SimpleAI-Detector and Fine-tuned DeBERTa maintain good performance. Specifically, \textbf{around 2 to 6 characters editing of typos or homoglyph per text can degrade the performance of most detectors to be worse than random} (The average length of texts is around 120 tokens).
All metric-based methods show a continuous decrease while attack budgets increase. In typo and homoglyph attacks, the decrease can mount up to a total failure with ROC AUC near 0.
In comparison, the DetectGPT is more robust than the others. \eg{} its drop converges to about 0.5 under homoglyph alteration while GLTR and Log Rank are near 0.
But, DetectGPT with fewer perturbed samples (\texttt{\_1d}) is more robust than larger ones (\texttt{\_10d} and \texttt{\_10z}), which is counterintuitive.

Among fine-tuned detectors, OpenAI Detectors show a similar unsoundness as metric-based ones. And their drops are most significant in format character editing.
SimpleAI Detector and F.t.-DeBERTa show great robustness to all editing attacks. 
Moreover,  comparing two underperforming OpenAI Detectors of different sizes indicates that larger classification models could be more robust than smaller ones under editing.
The watermarking method is also very robust to attacks, \ie{} keeping the AUC ROC above 99\%.

In addition, all four individual types of typo share similar negative impacts as the mixed version, as shown in \appref{app:detail_typo}. Also, all format character editing observed similar drops, while zero-width whitespace is the most effective.

\begin{figure}[t]
  \centering
  \includegraphics[width=1.0\linewidth]{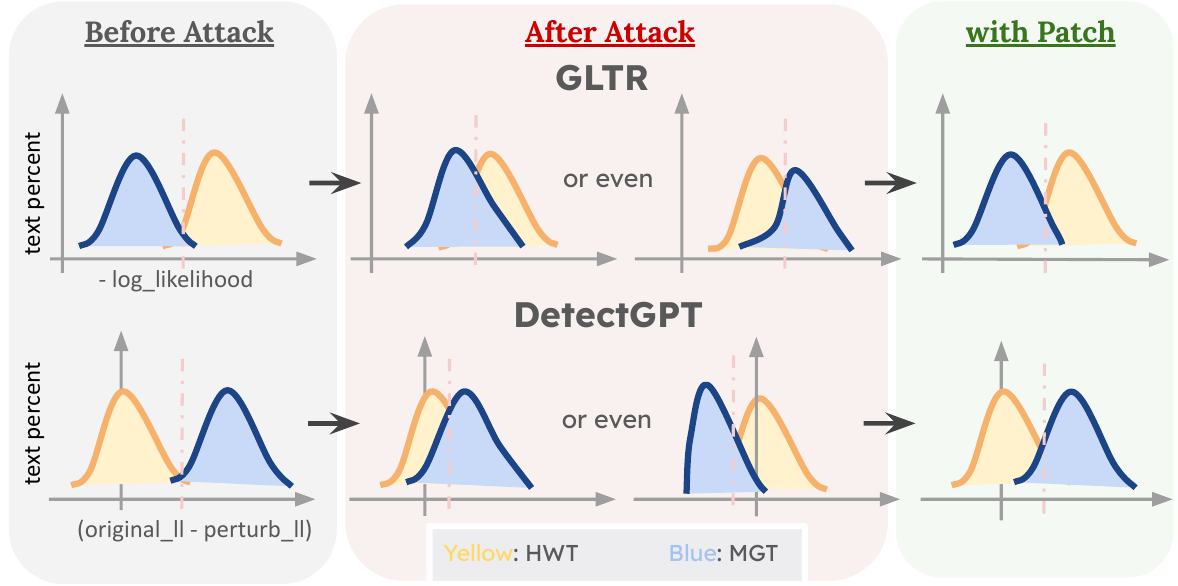}
  \caption{Illustration of the distribution of the metric value of the metric-based detectors before the attack, after the attack, and after patching (an out-of-the-box defense we proposed in \secref{patch}). The red dotted lines are the optimal decision boundaries.}
  \label{fig:patch} 
\end{figure}

\fakeparagraph{Interpretation.} The assumption of metric-based detectors is that HWTs have smaller log-probabilities than MGTs when inferenced by the generator model. 
However, editing attacks can effectively decrease the next-token probabilities, leading to indistinguishable situations of the distribution curves and even inverse relative relationships, which cause completely wrong predictions (ROC AUC near 0) as the budget increases.
\figref{fig:patch} shows a detailed illustration, taking GLTR and DetectGPT as examples. 
In the figure, after the attack, a larger overlap of the two curves (column 2) means more severe indistinguishability, and the interchange of the relative positions of the two curves (column 3) leads to wrong predictions. 
From this intuition, we attempt to patch the issue in \secref{patch} by removing anomalies.

\begin{figure*}[h]
\centering
\includegraphics[width=\textwidth]{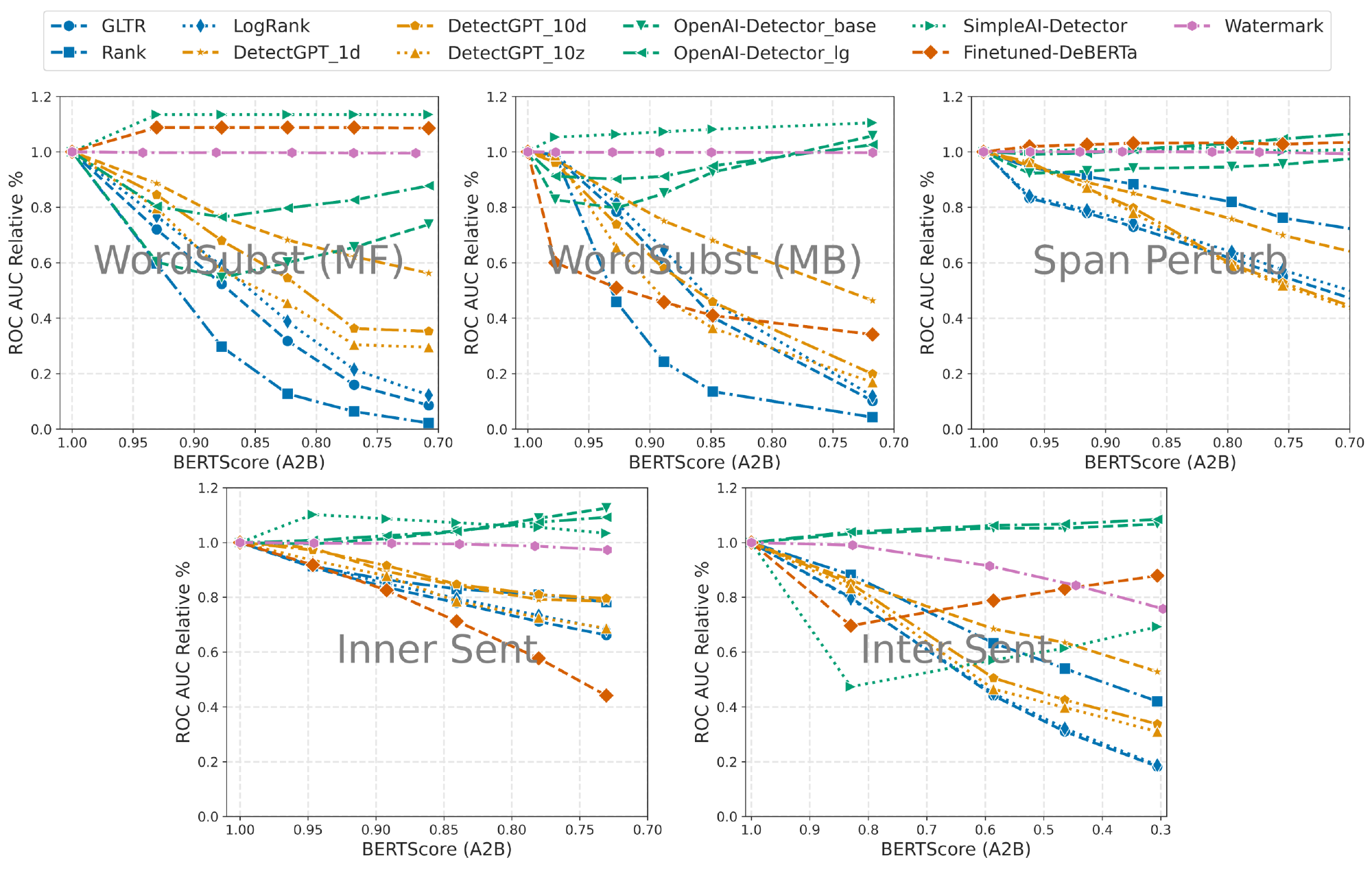}
\caption{\textbf{Performance drop of the detectors under the paraphrasing attacks.} We use BERTScore (A2B) as the budget in the figure.`A2B' means we compute BERTScore between unattacked MGTs and attacked MGTs. $\downarrow$ Smaller BERTScore value means a larger budget on the attack. Note that Inter Sent (sub-figure 5)'s x-limit interval is twice the previous.}
\label{fig:paraphrasing}
\end{figure*}

For the fine-tuned detectors, OpenAI Detectors perform worse in most cases, while SimpleAI Detector and F.t.-DeBERTa show great robustness. We surmise the reason is that OpenAI Detectors is trained on the GPT-2 corpus, which is outdated compared to the ChatGPT corpus for SimpleAI Detector and the GPT-J corpus for F.t.-DeBERTa. Under such an out-of-distribution (OOD) situation, the model shows less robustness. 

The watermarked detector adds a signature at each token, and our editing attacks only change a minimal portion of them. Hence, they show substantial robustness, maintaining high AUC ROC.


\subsubsection{Out-of-the-box Defense Patch\label{patch}}

\begin{table}[t]
\centering
\resizebox{0.85\linewidth}{!}{
\begin{tabular}{c c c c}
\toprule
\textbf{\texttt{Detector}} & \textbf{Before Att.} & \textbf{After Att.} & \textbf{w/ Patch} \\
\midrule
\textit{\textbf{DetectGPT-1d}} & 0.6866& 0.4299 & 0.5111\\
\textit{\textbf{DetectGPT-10d}} & 0.8312& 0.3301 & 0.6048\\
\textit{\textbf{DetectGPT-10z}} & 0.8516& 0.2735 & 0.6032\\
\bottomrule
\end{tabular}
}
\caption{Performance of DetectGPT after patching under typo insertion attack in terms of AUC ROC.}
\label{tab:patch-edit}
\end{table}

In this section, we propose a simple patch for the under-performing DetectGPT approach.
As the editing attacks mainly cause extremely low token probabilities to deceive the classification, we view them as anomaly points to filter them out.
Specifically, for each text, the top k\% tokens with the lowest probabilities would be prevented from being masked and perturbed when doing mask-filling. Next, we do not take their token probability into the computation. \tabref{tab:patch-edit} show the patch recovers performance by 0.2285 on average for 3 settings.

Other potential patches include adversarial training \citep{goodfellow2014explaining}, visual character embeddings \citep{Wehrmann2019LanguageAgnosticVE} for homoglyph, and preprocessing with grammatical error correction \citep{Bryant2022GrammaticalEC}. These approaches are more costly, and we leave them to future work.

\subsection{Paraphrasing Attacks\label{paraphrasing}}

Paraphrasing attacks aim to rewrite the generated texts without changing the semantic meanings at the post-generation stage.
Paraphrasing has been used for robustness evaluation and data augmentation in many other tasks, \eg{} sentiment analysis,
textual entailment \citep{iyyer-etal-2018-adversarial}, and machine translation \citep{Merkhofer2022PracticalAO}.
Usually, an extra LLM is used as the paraphraser \citep{iyyer-etal-2018-adversarial, yang-etal-2022-gcpg}. 
\citet{krishna2023paraphrasing} has reported the attack success of their paragraph-level paraphraser on some MGT detectors, but a comprehensive study across a wider range of paraphrasers and detectors is missing in the literature.
In this section, we will introduce five attack types that cover paraphrasing attacks of different granularity, from word-level to paragraph-level.
\subsubsection{Approaches}

\fakeparagraph{Synonyms Substitution\label{synonyms}}
is to replace some words with their synonyms to perturb the textual features.
Inspired by the red teaming setting of \citet{shi2023red}, we design a \textit{model-free} method and a \textit{model-based} method.
For the model-free substitution, we replace the selected words with their synonyms retrieved from a static dictionary WordNet \citep{miller-1994-wordnet}.\footnote{We avoid substituting the pronouns and prepositions to avoid grammatical problems. However, issues like verb tense still might happen.}
However, it does not consider the context of the substituted words. In the model-based method, we use T5-large \citep{raffel2020exploring} to select the words to be substituted and prompt LlaMA \citep{touvron2023llama} to get the synonyms given the context (detailed in \appref{app:synonym}).

\fakeparagraph{Span Perturbation\label{span_ptb}}
is to rewrite word spans like phrases or clauses. Compared to synonym substitution, span perturbation is more flexible in that tokens can be reordered or replaced.
Following the perturbation method of DetectGPT, we first randomly select spans for masking and then use T5-large to fill in. We control the length of the masked span but do not limit the length of tokens to be filled in.

\fakeparagraph{Inner-Sentence Paraphrase\label{inner_sent}} 
is to paraphrase each sentence separately. 
We use Pegasus \citep{zhang2020pegasus} to process sentences of texts and join them back to the full texts. 
To control the budget of attack, we can adjust the portion of sentences to be paraphrased.

\fakeparagraph{Inter-Sentence Paraphrase\label{inter_sent}} uses Dipper \citep{krishna2023paraphrasing} to paraphrase the whole text at once, which can reorder, merge, and split multiple sentences. We control the lexical diversity and order diversity to change the budgets. 

For budget, we measure the semantic difference between before- and after-attack with BERTScore.


\subsubsection{Results and Analysis}
The result is shown in \figref{fig:paraphrasing}.
\textbf{Interestingly, lower-level perturbations (\ie{} word substitution) show greater attack success than higher-level perturbations (\ie{} sentence-level paraphrases) at the same budget.}
Metric-based detectors show weakness at all level perturbations, especially degrading to near 0.0 AUC ROC under word substitution attacks. 
Among metric-based detectors, DetectGPT shows slightly better robustness at word-level perturbation but then loses the lead at higher levels.
For fine-tuned detectors, SimpleAI Detector remains robust under all attacks, while OpenAI Detectors and F.t.-DeBERTa fail at some attacks. 
A surprising result is that in some cases, fine-tuned detectors' performance first drops but then increases as the budget increases, \eg{} OpenAI Detectors under word substitution and F.t.-DeBERTa under inter-sentence paraphrase.
Finally, \textbf{for watermarking, inter-sentence paraphrasing is the only attack effective.} 

\begin{table}[t]
\centering
\resizebox{\linewidth}{!}{
\renewcommand\arraystretch{1.2}
\begin{tabular}{l l c c c c }    
\toprule
\textbf{AUC\%} & \textbf{\texttt{Attack}} & \multicolumn{2}{c}{\textbf{\textit{P.-Para}}}& \textbf{\textit{ICL}} & \textbf{\textit{CS Gen}} \\
& \textbf{Dataset}& \textit{GPT-J}& \textit{GPT-4}&\textit{GPT-4}&\textit{GPT-4}\\
\midrule
& PPL unatt.& 1.930 & 2.042 & 2.042 & 2.042\\
& MAUVE unatt.& 0.944 & 0.483 & 0.483 & 0.483\\
\midrule
\multirow{2}{*}{\textbf{\texttt{Budget}}}& PPL attacked& 1.867 &2.064& 2.080 &  4.971\\
& MAUVE att.& 0.963 & 0.348 & 0.680 & 0.056 \\
\midrule
\multirow{12}{*}{\textbf{\texttt{Detect.}}} & GLTR & 105.3  &111.3&  96.83
&  \cellcolor[HTML]{EC7063} \textbf{16.40}
\\
 & Rank & 103.8  &114.5&  95.15
&  \cellcolor[HTML]{EC7063} \textbf{13.47}
\\
 & LogRank & 105.0  &111.7&  97.37
&  \cellcolor[HTML]{EC7063} \textbf{16.58}
\\
 & DetectGPT-1d & 99.64  &109.4&  98.96
&  \cellcolor[HTML]{F5CBA7 } \textbf{59.68}
\\
 & DetectGPT-10d & 99.98  &112.9&  96.76
&  \cellcolor[HTML]{EC7063} \textbf{31.44}
\\
 & DetectGPT-10z & 99.94  &112.9&  97.15
&  \cellcolor[HTML]{EC7063} \textbf{35.62}
\\
 & OpenAI Det.-Bs & 115.9  &135.8&  96.71
&  \cellcolor[HTML]{F5CBA7 } \textbf{54.04}
\\
 & OpenAI Det.-Lg & 110.4  &128.1&  99.79
&  \cellcolor[HTML]{F5CBA7 } \textbf{57.25}
\\
 & SimpleAI Det. & \cellcolor[HTML]{EC7063} \textbf{25.63}  &\cellcolor[HTML]{EC7063} \textbf{33.20}&  102.64
&  107.44
\\
 & F.t. DeBERTa & \cellcolor[HTML]{EC7063} \textbf{43.70}  &98.19
&  99.70
&  108.13
\\
 & Watermark* & 99.98  & - - & - - & - - \\
\bottomrule
\end{tabular}
}
\caption{\textbf{Performance drop of the detectors under the prompting attacks.} The perplexity (abbr. PPL) and MAUVE (M2H) are as the budgets for quality. For short, `P.-Para' is the prompt paraphrasing attack, and `CS Gen' is the character-substituted generation attack. 
Bolding indicates severe performance drop (drops larger than 50\% are in red; between 25\% and 50\% are in yellow).\footnotemark \label{tab:prompt-att}}
\end{table}
\footnotetext{Due to GPT-4 being close-sourced, we can not test the watermark on it.}

\fakeparagraph{Interpretation.}
For metric-based detectors, localized disturbances from lower-level perturbations cause more decreases in next-token probability than high-level perturbations. While for high-level perturbations, the decrease is spread out in wider spans, thus minor the overall impact.
For fine-tuned detectors, \citet{liu2022coco} pose that they concentrate more on long-form patterns (\eg{} commonly used phrases or sentence structures) from LLM to detect. Hence, localized disturbances of low-level perturbation directly interrupt the long-form patterns, while high-level paraphrasing is milder as it rewrites such patterns but still keeps some of the machine signatures.
Moreover, we surmise that paraphrasing attacks are not making MGTs more human-like but only mixing the machine signatures. So, sometimes, the detectors' performance falls then rises as the budgets increase, during which the dominant machine signatures switch from the original generator's to the paraphraser's.

Discussion on future defense is deferred to \appref{app:para_defense}.


\subsection{Prompting Attacks\label{prompting}}

Most detectors are trained and tested on data based on fixed, well-designed prompts, \eg{} question answering \citep{guo2023close}, translation \citep{su2023hc3}, continually writing \citep{zellers2019defending}, etc{}. But in realistic scenarios, user prompts might be much more diverse, abnormal, and even noisy \citep{ZamfirescuPereira2023WhyJC}.
In this section, we will introduce three attack types and their impact on detectors, calling for more inclusive designs for detectors.
All prompting attacks are in the pre-generation stage.

\subsubsection{Approaches}

\begin{table}[t]
\centering
\resizebox{0.9\linewidth}{!}{
    \begin{tabular}{l L{7cm}}
    \toprule   
         \small \textbf{\textit{Prompt:}}& {\begin{tabular}{@{}L{7cm}@{}}\small Continue 20 words with all `a's substituted with `z's and all `z's substituted with `a's: \\
         \small As the sun dipped below the horizon, casting\end{tabular}}  \\
    \midrule
         \small \textbf{\textit{\color{magenta}{GPT-4:}}}&  \small Zs the sun dipped below the horiaon, czsting shzdows zcross the lzndsczpe, z gentle breeae whispered through the trees, czrrying with it the sweet zromz of spring flowers ...\\
    \midrule
          {\begin{tabular}{@{}l@{}}\small \textbf{\textit{\color{teal}{Cleaned}}} \\ \small \textbf{\textit{\color{teal}{Output:}}}\end{tabular}} &  \small As the sun dipped below the horizon, casting shadows across the landscape, a gentle breeze whispered through the trees, carrying with it the sweet aroma of spring flowers ...\\
    \bottomrule
    \end{tabular}
    
}
    \caption{A character-substituted generation example.}
    \label{tab:cs-gen-example}

\end{table}

\fakeparagraph{Prompt Paraphrasing\label{para_prompt}}. Instead of paraphrasing whole texts post-generation (\secref{paraphrasing}), paraphrasing the prompt prior to generation is much cheaper and low-impact on the output quality. 
We use Pegasus paraphraser to rewrite the prompts.\footnote{Since the prompts are usually very short, it is hard for us to control the budget while paraphrasing. Hence, we report attacked performance under a single budget at \tabref{tab:prompt-att}.}

\fakeparagraph{In-Context Learning\label{icl}} \citep{dong2022survey} improves generation quality by giving only a few examples in the form of demonstration. 
To generate more human-like texts to deceive detectors, we give the generator a related HWT as a positive example and a vanilla MGT as a negative example. We follow the prompt design of Super-NaturalInstructions \citep{wang2022super}.\footnote{It is also hard to adjust the budget for this attack. One potential way is to change the demonstration number, but it shows no clear correlation to the budgets and also might exceed the generator's maximum length of the input sequence.}

\begin{figure*}[h]
\centering
\includegraphics[width=0.7\textwidth]{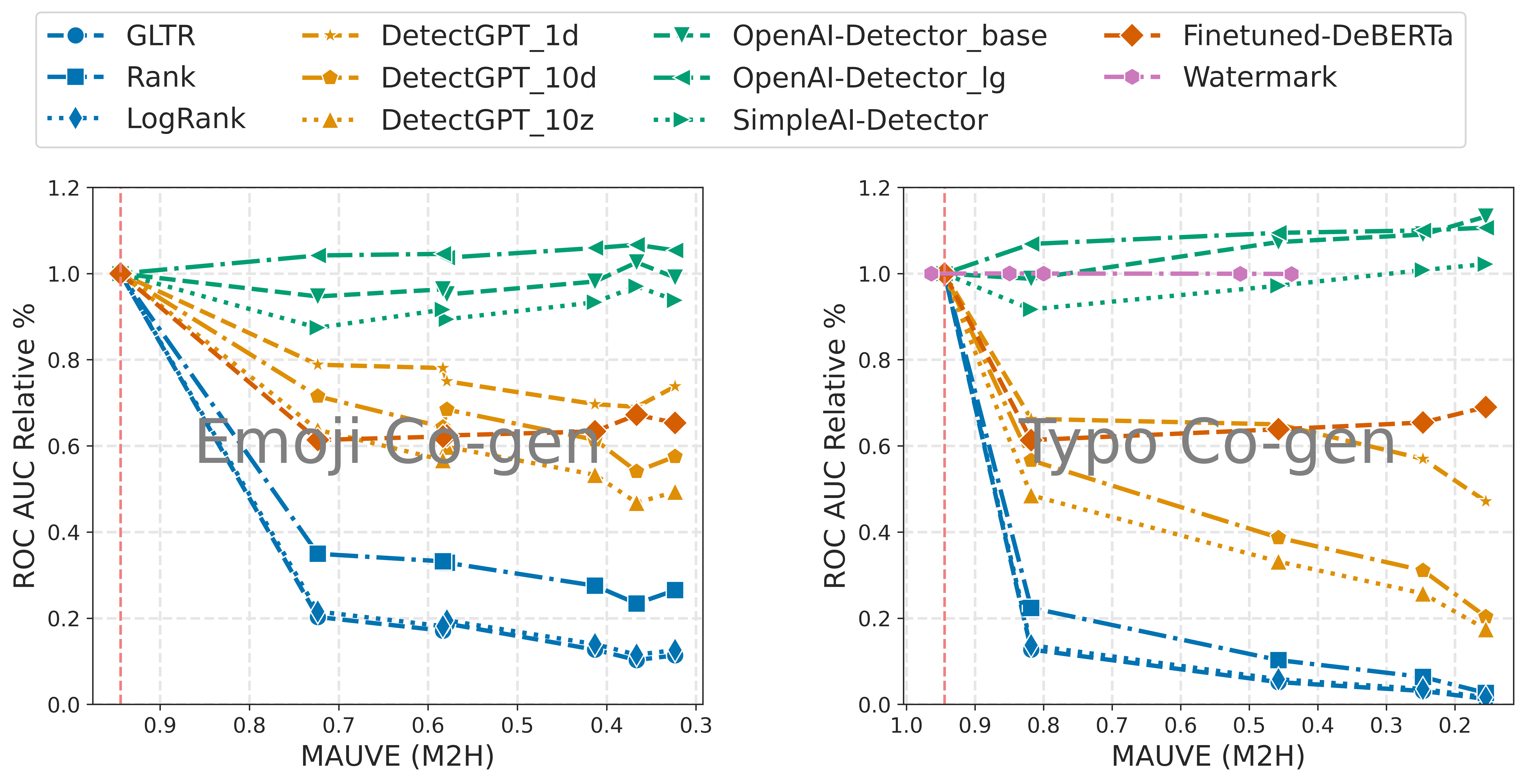}
\caption{\textbf{Performance drop of the detectors under the co-generation attacks.} We use MAUVE (M2H) as the budget to evaluate the text quality in the figure.`M2H' means we compute MAUVE between HWTs and attacked MGTs. The vertically dotted red line is the score w.o. attack. $\downarrow$ Smaller MAUVE (M2H) value means a larger budget on the attack.}
\label{fig:cogen}
\end{figure*}

\fakeparagraph{Character-Substituted Generation\label{char_sub}} is inspired by substitution cipher \citep{Spillman1993UseOA} in the domain of cryptography. We found that recent LLM, \eg{} GPT-4, are able to follow some easy substitution rule, \eg{} \texttt{replace `a's with `z's}, during generation. Afterward, we substitute the mapping back to clean the output. \tabref{tab:cs-gen-example} shows an example.
Notably, we only find GPT-4 maintains acceptable generation quality under this attack, so we solely show its result.


\subsubsection{Results and Analysis}

\tabref{tab:prompt-att} shows the results. The metric-based detectors severely suffer the character-substituted generation attack but are robust to prompt paraphrasing and in-context learning.
On the other hand, the fine-tuned detectors show some drop in prompt paraphrasing but have great robustness under character-substituted generation.\footnote{Note that the budget of character-substituted generation is larger than other attacks. As a prompting method, it is hard to control it, so a milder character-substitution method with an adjustable budget is by controlled generation (\secref{co-gen}).}

\fakeparagraph{Interpretation.} The character-substituted generation attack is a more localized perturbation compared with prompt paraphrasing and in-context learning, which is on the general level.
So, similar to the paraphrasing attacks, metric-based detectors show a larger vulnerability to localized perturbation since it directly increases the next-token probabilities, which is also shown as the high perplexity after the attack.
However, fine-tuned detectors focus more on long-term patterns, which may not impacted by a few substitutions. But, prompt paraphrasing is a form of attack that shifts the prompt pattern, which can degrade fine-tuned detectors severely, especially those ones that are not generalizable.  

Discussion on future defense is deferred to \appref{app:prompt_defense}.


\subsection{Co-Generating Attacks\label{co-gen}}

Instead of directly perturbing the prompt, we propose \textit{co-generating}, which perturbs the generated tokens at each recurrent step with some designed rules.
The mechanism of co-generating attacks shares similarities to the typo insertion attack (\secref{editing}). But for co-generating, the perturbed text is cleaned to be grammatically correct after generation. 
We will introduce two attack types: emoji and typo co-generation. 

\subsubsection{Approaches}

\fakeparagraph{Typo Co-Generation\label{sec:typo-co-gen}} is to insert typos during generation.  
Different from the typo insertion attack (\secref{editing}), we introduce typos immediately after the token is sampled (before the generation of the next token), following preset typo insertion rules, \eg{} substitute all `a's into `z's. After the whole generation is finished, we reverse the inserted typos to clear the errors.
Compared to the typo insertion attack, typo co-generation does not directly damage quality and human imperceptibility.

\fakeparagraph{Emoji Co-Generation} is developed in a similar fashion: We insert emojis at the end of generated sentences (before the generation of the next sentences) and remove them post-generation. The details are deferred to \appref{app:emoji-co-gen}. 

\subsubsection{Results and Analysis}

\figref{fig:cogen} shows the results. We observe that the metric-based detectors and F.t.-DeBERTa are not robust to the co-generation attacks, while OpenAI and SimpleAI Detectors show minor degradation. 
DetectGPT is more robust than other metric-based methods without perturbation, \eg{} it converges at around 0.5 under typo co-generation while GLTR and (Log-)Rank converge near 0.1.
For all detectors, the further increase in budgets for co-generation attacks does not cause proportional performance drops.

\fakeparagraph{Interpretation.} The insertion of emojis and typos during recurrent next-token generation is a disruption for the sampling of LLMs, shifting the generation away from the generator's original distribution. 
Moreover, removing the emojis and recovering the typos post-generation disrupt the conditional probability again for metric-based detectors.
For fine-tuned models, we surmise that when doing in-domain detection (F.t.-DeBERTa), the detector might focus more on localized features. Otherwise, out-of-domain models here (OpenAI and SimpleAI Detectors) focus on long-term patterns. 
Thus, the in-domain model is less robust to the attacks.

Discussion on future defense is deferred to \appref{app:cogen_defense}.

\section{Related Work}

To the best of our knowledge, there is no existing thorough study on the robustness of machine-generated text detection under various attacks. The most related studies are:

\fakeparagraph{Study on Adversarial Attack to MGT Detection.}
Adversarial attack \citep{goodfellow2014explaining}, which exposes optimized regions of the input space where the model under-performs, first introduced to text data by \citet{li2018textbugger}, is powerful to reveal robustness in text classification \citep{jin2020bert}. 
\citet{shi2023red} first use adversarial attack on MGT detectors, including OpenAI-Detector, DetectGPT, and watermarking. They cover adversarial word substitution and adversarial prompting, both of which deceive three detectors.
Furthermore, RADAR \citep{hu2023radar} attempts to improve the robustness of model-based detectors by adversarial learning on paraphrasing. We also take inspiration from recent work on the blind spots of NLG metrics \citep{he-etal-2023-blind}.

However, under realistic scenarios, attackers do not have detailed knowledge of which detector is being used (\secref{sec:scope}). Our work focuses on non-adversarial attacks, which are less costly and under-explored.


\fakeparagraph{Study on Generalization of MGT Detection.}
Generalization capability is an important aspect of robustness in MGT detection.
For model-based detectors, \citet{solaiman2019release} evaluate their OpenAI Detector on generalize through different model sizes, sampling strategies, and input text length. 
\citet{pu2023zero} study the generalization ability when training and testing on data from different generators.
\citet{pagnoni-etal-2022-threat} analyze the generalization on sequence length, decoding strategy, dataset domain, and generator size.
\citet{wang2023m4} introduce a multi-generator, multi-domain, and multi-lingual corpus to train more generalizable detectors.
For metric-based detectors, \citet{mireshghallah2023smaller} explore the generalization between different base models and dataset generators on a perturbation-based metric-based detector.
In comparison, our research focus is not on the generalization problem but on the robustness against realistic and malicious attacks.

\fakeparagraph{Robustness of MGT Detection.}
Some existing works of MGT detectors evaluate their robustness under some specific attacks.
\citet{liu2022coco} evaluate the robustness of their model-based detector CoCo under token editing.
\citet{krishna2023paraphrasing} stress test detectors on paragraph-level paraphrase, and further purpose a retrieval-based method to increase robustness. 
\citet{hu2023radar} focus on paraphrastic robust model-based detectors by adopting adversarial learning.
\citet{zhang2023assaying} purpose that topic shifting drops the metric-based detectors' performance.
In the watermark domain, \citet{kirchenbauer2023watermark} propose a list of initial attack ideas, including editing, paraphrasing, and generation strategy. But, they only experiment on the span perturbation attack for their watermark method. 
Further, \citet{kirchenbauer2023reliability} study the watermark robustness after LLM paraphrase, manual paraphrase, and mix into a longer document. 
\citet{kuditipudi2023robust} purpose a distortion-free watermark that is robust against perturbation.
\citet{zhao2023provable} enhance the robustness of the watermarking scheme against editing and paraphrasing attacks by employing a fixed group design.
And \citet{hou2023semstamp} propose a semantic watermark at the sentence level for paraphrastic robustness. 
To summarize, a thorough and comparative study on the robustness covering a wide range of detectors and attacks is lacking in the literature, which motivates our work.

\section{Conclusion}
\textbf{Due to lack of space, we defer discussions on future work for attacks and defenses to \appref{app:future_work_a} and \appref{app:future_work_d}.}

This study evaluates the robustness of 8 MGT detectors against 12 realistic attacks, revealing striking vulnerabilities. 
Findings show that no detector consistently withstands all attacks, as some attack strategies severely compromise detection accuracy. 
Among various detectors, watermarking emerges as the most robust approach, followed by model-based detectors. 
We also suggest combining metric- and model-based detectors for better resilience. 
Aiming at robust MGT detection, we call for awareness of diverse attacks and highlight the need for more robust methods. 

\section*{Limitations}

We mainly show and discuss the results of representative generators, detectors, and attack methods in the main paper following the preset scope \secref{sec:scope}. Since our work is a general and reproducible evaluation pipeline, it is readily applicable to other generators or detectors. 


We mainly focus on English in our work. Most attacks are able to be generalized to other languages, but the generation quality might suffer mainly depending on the generator's capability, especially in lower-resource languages. Also, the detection accuracy highly relies on the base model's capability in other languages.
Some attacks could have slightly different designs for other languages, \eg{} the homoglyph alteration attack could be more complex in logographic languages like Chinese, Japanese (Kanji), and Vietnamese (Chu Nôm), and it would be interesting to explore in future work.

\section*{Ethics Statement}
The goal of this paper is not to provide a cookbook for malicious use of attacks to deceive MGT detectors.
On the contrary, we want to draw attention to the potential vulnerabilities of current MGT detectors. 
Moreover, we call for future MGT detectors that are robust against the attacks we tested. For this target, we will open-source all the code and dataset for easy reproduction of our pipeline of robustness tests. We also propose and describe some defense patches for fixing these loopholes.


\bibliographystyle{acl_natbib}

\appendix




\newpage

\newpage

\section{Experiment Settings}

The experiments are done on 8 Tesla V100 and 4 Tesla A100 GPUs, taking up a total of around 500 GPU hours.

\subsection{Dataset and Generators\label{app:dataset_gen}}

We build the dataset based on \citet{pu2023zero}. The HWTs are from the News domain of the dataset, and the MGTs are generated with different temperatures for each generator we selected.
\tabref{tab:sample_number} shows the sample number of each split in our dataset.

\begin{table}[h]
    \centering
    \begin{tabular}{cccc}
    \toprule
         Split&  Train&  Eval& Test\\
         \midrule
         Sample Num. &  8,000&  1,000& 1,000\\
    \bottomrule
    \end{tabular}
    \caption{The sample number of each split of the dataset.}
    \label{tab:sample_number}
\end{table}

For sampling, we use a combination of nucleus sampling \citep{welleck2019neural} with top-p = 0.96 and a tuned temperature parameter \citep{Caccia2020Language, nadeem-etal-2020-systematic}.
While smaller temperature gives higher quality, it will also cause repetition, especially for less capable LMs.
So, we tune the temperature based on the criteria of preventing repetition, which is < 0.2 in terms of 4-gram duplication under metric seq-rep-4 in \citet{welleck2019neural}.
\tabref{tab:temp} shows the our temperature settings.


\begin{table}[h]
    \centering
    \resizebox{\linewidth}{!}{
    \begin{tabular}{ccccccc}
    \toprule
         \texttt{Generator}&  GPT-2 XL&  GPT-J& LlaMA & LlaMA-2 &  DaVinci-003&GPT-4\\
         \midrule
         Temp.&  1.5&  1.5& 1.0 & 1.5 &  0.7&0.7\\
    \bottomrule
    \end{tabular}
    }
    \caption{The temperature we set for each generator to follow the criteria of avoiding severe repetitions.}
    \label{tab:temp}
\end{table}

\subsection{Detectors}

\subsubsection{Detail Hyperparameters\label{app:detail-hyper}}

For all model-based detectors, we use the original generator of the test set as the base model to compute the next-token probability and perplexity.

For \textit{DetectGPT}, we follow the recommendation hyperparameter setting. The perturbation word ratio is 15\% on 2-spam, the perturbation model is T5-3B \citep{raffel2020exploring}, and the sample number of perturbation is 1 or 10 (indicated in the name of the legend). In the legend, mode `d' represents the direct use of the absolute likelihood drop while mode `z' adds an additional normalization. The mask-filling in perturbation is with temperature 1 without any sampling strategy (\eg{} top-p and top-k).

For all fine-tuned detectors, we directly use the logits as the output probability. When fine-tuning \textit{F.t. DeBERTa}, we set batch size as 4, learning rate as 1e-5, weight decay as 0, adam epsilon as 1e-8, and epoch number as 10.

For the watermark, we follow the setting in \citet{kirchenbauer2023watermark}, setting gamma as 0.25, seeding scheme as selfhash, and z-score threshold as 4.0.

\subsubsection{Under Closed-Source Dataset}

For GPT-4 datasets, as we do not have the white-box generator model, we select an alternative LM as the base model. 
According to the conclusion from \citet{mireshghallah2023smaller}, GPT-2 Small \citep{Radford2019LanguageMA} is the best-performed base model when generalized to GPT-4. Our experiment compares GPT-2 (Small, Medium, Large, XL), OPT (125M, 350M, 1.3B, 2.7B) \citep{zhang2022opt}, GPT-Neo (125M, 1.3B, 2.7B) \citep{gpt-neo}, and GPT-J (6B), and the results align that GPT-2 Small is the best.
Hence, our GPT-4 dataset results are all under GPT-2 Small as the base model. \tabref{tab:unattacked-gpt4} shows the unattacked performance. \tabref{tab:base_budget-gpt-4} shows the Perplexity and MAUVE (M2H) as budget of unattacked GPT-4 dataset.

\begin{table}[h]
\centering
\resizebox{0.6\linewidth}{!}{
    \begin{tabular}{cc}
        \toprule
         \multicolumn{2}{c}{\textbf{\textit{Unwatermarked}}}\\
         \midrule
         \textit{PPL}&  \textit{MAUVE(M2H)}\\
         2.042 $\pm$ 0.250&  0.4831\\
         \bottomrule
    \end{tabular}
    }
    \caption{The unattacked value of average Perplexity and MAUVE (M2H) of GPT-4 dataset as the base point budget. }
    \label{tab:base_budget-gpt-4}
\end{table}

\begin{table}[h]
\centering
\resizebox{\linewidth}{!}{
\renewcommand\arraystretch{1.2}
\begin{tabular}{l c c c c c }  
\toprule
\texttt{\textbf{Detector}} & \textit{AUC} & \textit{TF=5} & \textit{TF=10} & \textit{TF=20} & \textit{ACC} \\
\midrule
GLTR & 62.41& 2.20& 7.20& 22.00& 60.40
\\
Rank & 62.15& 11.40& 21.20& 36.00& 59.80
\\
LogRank & 65.96& 5.00& 17.00& 32.60& 62.00
\\
Entropy & 66.40& 12.40& 23.00& 35.80& 61.80
\\
\midrule
DetectGPT-1d & 51.22& 3.60& 6.60& 15.20& 50.40
\\
DetectGPT-10d & 55.61& 2.00& 4.40& 16.20& 55.60
\\
DetectGPT-10z & 59.53& 5.80& 11.00& 22.20& 58.40
\\
\midrule
OpenAI Det.-Bs & 55.72& 13.20& 20.20& 28.60& 52.60
\\
OpenAI Det.-Lg & 57.70& 6.00& 12.20& 24.60& 56.00
\\
SimpleAI Det. & 86.81& 81.00& 82.20& 85.40& 84.40
\\
F.t. DeBERTa & 100.0& 99.80& 99.80& 99.80& 99.80
\\
\bottomrule
\end{tabular}
}
\caption{\textbf{The performance of detectors in the unattacked scenario for the GPT-4 dataset.} For short, `AUC' is ROC AUC, `TF=5' is TPR@FPR=5\%, `ACC' is Accuracy, `Det.' is Detector, and `F.t.' is Fine-tuned.}
\label{tab:unattacked-gpt4}
\end{table}

\section{Future Work \label{app:future_work}}

\subsection{Future Work on Defenses \label{app:future_work_d}}

\subsubsection{Paraphrasing Attacks (\secref{paraphrasing}) \label{app:para_defense}}

For metric-based detectors, a straightforward way is to choose a base model that is related to common paraphrasers' base models, \eg{} T5, ProphetNet \citep{qi2020prophetnet}, or fine-tune the base model on some paraphrased corpus. 
Similarly, data augmentation on paraphrasing and adversarial learning could be useful for training fine-tuned detectors \citep{hu2023radar}.
Moreover, \citet{krishna2023paraphrasing} purpose that retrieval on an MGT database can be the defense, if it is possible to collect enough in-domain MGT entries. 
For watermarking, semantic-level watermarking \citep{hou2023semstamp} (as opposed to token-level) has been proposed for paraphrastic robustness.

\subsubsection{Prompting Attacks (\secref{prompting}) \label{app:prompt_defense}}

To patch the weakness of fine-tuned detectors under prompt paraphrasing, an efficient way is to fine-tune the classifier on multi-generator, multi-domain datasets, \eg{} M4 by \citet{wang2023m4}. Otherwise, using an ensembling system \citep{pagnoni-etal-2022-threat} containing both metric-based and fine-tuned detectors could ease the problem.
However, we surmise there is no direct way to patch the character-substituted generation because it mimics the suboptimal generation strategy of humans at the root. Yet current LLMs are not capable of always following the character-substitution prompts with high text quality, which could cause unnatural expressions and extra typos.
A way to fix the loophole could be censoring the prompts and generated texts if they have weird expressions \citep{dou-etal-2022-gpt, chiang-lee-2023-large}. Additionally, training detectors on the MGT corpus from unnatural instructions \citep{honovich2022unnatural} could also be considered.

\subsubsection{Co-Generating Attacks (\secref{co-gen}) \label{app:cogen_defense}}

To the best of our knowledge, there are no related existing works. We feel it is hard for metric-based detectors to overcome the defects. Under this scenario, fine-tuned detectors could be the better choice. One potential way to enhance fine-tuning is to adopt some data augmentation, like random masking on short-term spans. 
Also, we surmise a combination of metric-based detectors and model-based detectors is useful to bypass each other's stumbling blocks better when attacked. The ensembling could also ease the impact of other attacks. 
Fortunately, the co-generation attacks are still not widely available now since they need to be on the white-box models.

\subsection{Future Work on Attacks \label{app:future_work_a}}

Below, we briefly discuss other types of attacks related to generalization, which are not covered in this work.

\fakeparagraph{Sampling Attacks.}
Diverse sampling strategies \citep{holtzman2019curious} can be adopted when generating MGTs both by setting different hyper-parameters. 
\citet{pagnoni-etal-2022-threat} show that detection performance generally decreases when a fine-tuned detector is evaluated on a sampling strategy it was not trained on. 

\fakeparagraph{Fine-Tuning Attacks.}
In some scenarios, users might fine-tune the generator LLM on their specific domain.
Since the detectors have no knowledge and access to the customized generator, their performance might decrease.

\fakeparagraph{Human-Involved Attacks\label{human-involved}}
is to manually polish or replenish MGTs to be more human-like and improve their quality, which could deceive the MGT detector.
\citet{kirchenbauer2023reliability} purpose manual paraphrasing and mixing HWTs into MGTs as an attack to watermarks. And \citet{christ2023undetectable} describe a manual prefix-specificity scheme to lead to a more human-like generation.
Therefore, a major limitation of the current detector technique is the inability to classify human-LLM-collaborated texts into binary classes.
Future MGT detectors that are able to measure the portion of LLM involvement in text writing are worth considering as an answer to this attack genre.

\section{Attack Details}

In this section, we report the details that are not included in the main paper due to lack of space, including methodologies and settings.

\subsection{Typo Insertion \label{app:detail_typo}}

\tabref{tab:app:4typo} shows the performance drop of four separate typo types, \ie{} insertion,
deletion, substitution, and transposition. All of them share similar observations on degradation trends and are close to the mixed typo type. Therefore, for the figure in the main text, we show the result of mixed for brevity.

\begin{table}
\centering
\resizebox{\linewidth}{!}{
\renewcommand\arraystretch{1.3}
\begin{tabular}{l l c c c c c}  
\toprule
\textbf{AUC\%} & \textbf{Typo Type} & \textbf{\textit{Mixed}} & \textbf{\textit{Insert}} & \textbf{\textit{Delete}} & \textbf{\textit{Subst.}} & \textbf{\textit{Trans.}} \\
 \midrule
Budget & Edit Distance & 17.68 & 18.05 & 18.04 & 16.76 & 17.87 \\
\midrule
\multirow{12}{*}{\texttt{\textbf{Detect.}}} & GLTR & 2.14 & 2.96 & 6.96 & 3.17 & 5.76 \\
 & Rank & 6.81 & 7.25 & 13.70 & 6.67 & 12.18 \\
 & LogRank & 2.56 & 3.65 & 9.74 & 3.72 & 7.67 \\
 & DetectGPT-1d & 44.66 & 44.57 & 53.38 & 42.59 & 58.28 \\
 & DetectGPT-10d & 17.99 & 15.98 & 32.62 & 18.01 & 25.18 \\
 & DetectGPT-10z & 15.02 & 14.54 & 26.24 & 15.95 & 20.75 \\
 & OpenAI Det.-Bs & 27.62 & 27.37 & 24.00 & 26.32 & 25.57 \\
 & OpenAI Det.-Lg & 34.76 & 29.56 & 35.11 & 32.68 & 33.58 \\
 & SimpleAI Det. & 111.6 & 111.1 & 112.1 & 111.0 & 111.2 \\
 & F.t. DeBERTa & 108.4 & 96.80 & 97.20 & 96.83 & 97.48 \\
 \bottomrule

\end{tabular}
}
\caption{Detectors' performance drops in terms of relative AUC ROC \% of 4 typo types, namely insertion, deletion, substitution, and transposition.}
\label{tab:app:4typo}
\end{table}

\subsection{Homoglyph Alteration\label{app:homoglyph}}

We consider \texttt{\textbackslash n} - newline, \texttt{\textbackslash r} - carriage return, \texttt{\textbackslash v} - vertical tab, \texttt{\textbackslash u200B} - zero-width whitespace, and \texttt{\textbackslash u000B} - line tabulation as representatives and they shared similar results.
Specifically, zero-width whitespace can be inserted between any tokens, while we only add shift-related characters at the end of sentences.

\subsection{Synonym Substitution \label{app:synonym}}

\tabref{tab:app:syn-prompt} shows the prompt design for LlaMA to do the model-based synonym generation with the context. After the generation, we have an additional step to ask LlaMA double check and correct the grammar of the substituted sentences.

\begin{table}[h]
\small
\begin{tabular}{M{0.95\linewidth}}
\toprule
\$\{sentence\}\textbackslash n Synonyms of the word ``\$\{word\}" in the above sentence are:\textbackslash n a)\\
\bottomrule
\end{tabular}
\caption{Prompt for LlaMA to generate synonyms based on the context for substitution attack.}
\label{tab:app:syn-prompt}
\end{table}

\subsection{Typo Co-Generation \label{app:co-gen}}

The results reported in the main text using the typo substitution rule switching `c's and `k's.
We have also tried other rules, \eg{} `a's and `z's. The different rules cause different budgets depending on the character appearance frequency in the texts. We select a rule that has a comparable budget interval to other attacks, but our system also supports other rules.

\subsection{Emoji Co-Generation \label{app:emoji-co-gen}}
Emojis are widely used in web texts, especially social media \citep{Ayvaz2017TheEO}.
However, emojis are usually excluded from the training corpus of fine-tuned detectors and are situated at the long tail of distribution for metric-based detectors. Thus, they have a similar effect as the insertion of typos (\secref{editing}).
We insert a random emoji from Gemoji\footnote{A package of emoji collections: \url{https://github.com/wooorm/gemoji}.} when LLM finishes a sentence and let the LLM generate the next sentence recurrently.
We control the budgets by tuning the probability of inserting an emoji after a sentence.
We clean the output texts after generation by removing all emojis to hide the trace of the attack. Note that the distribution shift caused by emoji during sampling will still embodied in the text and deceive the detectors.

We have also attempted emoji co-generation for watermarking, and it also demonstrates very strong robustness, similar to the typo case. Interestingly, inserting more emojis did not affect the budget (MAUVE score) for watermarked generation. Therefore, we choose not to plot this result in \tabref{fig:cogen} to avoid confusion.

\section{Additional Results\label{app:add_res}}

\subsection{Across Budgets\label{app:acs_budgets}}

The design of the budget considers the alignment of different metrics's indications, especially for the ones on the same aspects.

\begin{figure*}[t]
\centering
\includegraphics[width=\textwidth]{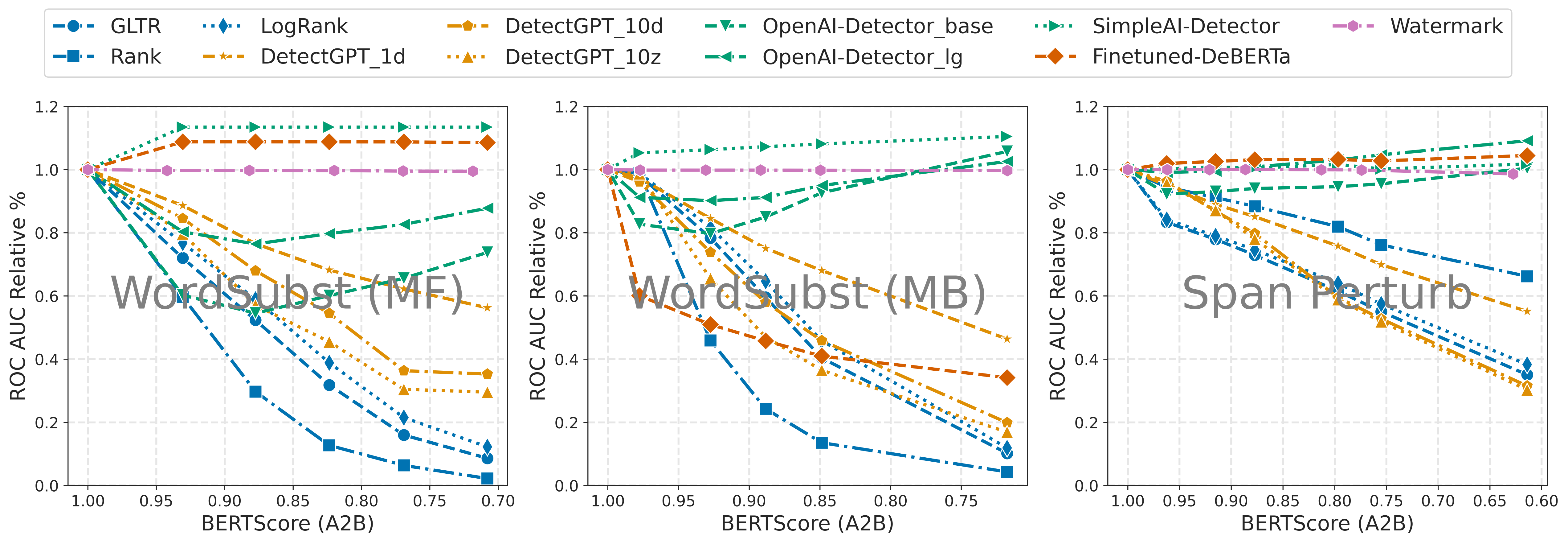}
\caption{Performance drop under the paraphrasing attacks with \textbf{BERTScore} (A2B) as budget (x-axis). (Row 1)}
\label{app:fig:para_upper.BERTScore}
\end{figure*}

\begin{figure*}[t]
\centering
\includegraphics[width=\textwidth]{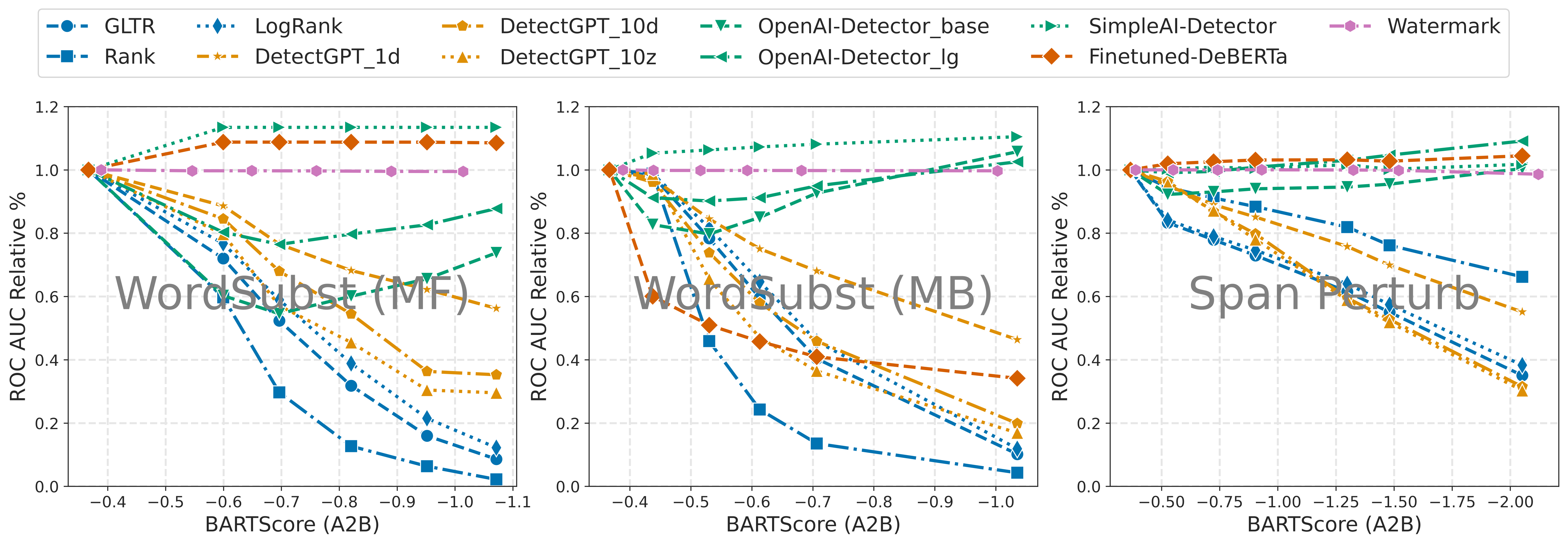}
\caption{Performance drop under the paraphrasing attacks with \textbf{BARTScore} (A2B) as budget (x-axis). (Row 1)}
\label{app:fig:para_upper.BARTScore}
\end{figure*}

\begin{figure*}[t]
\centering
\includegraphics[width=\textwidth]{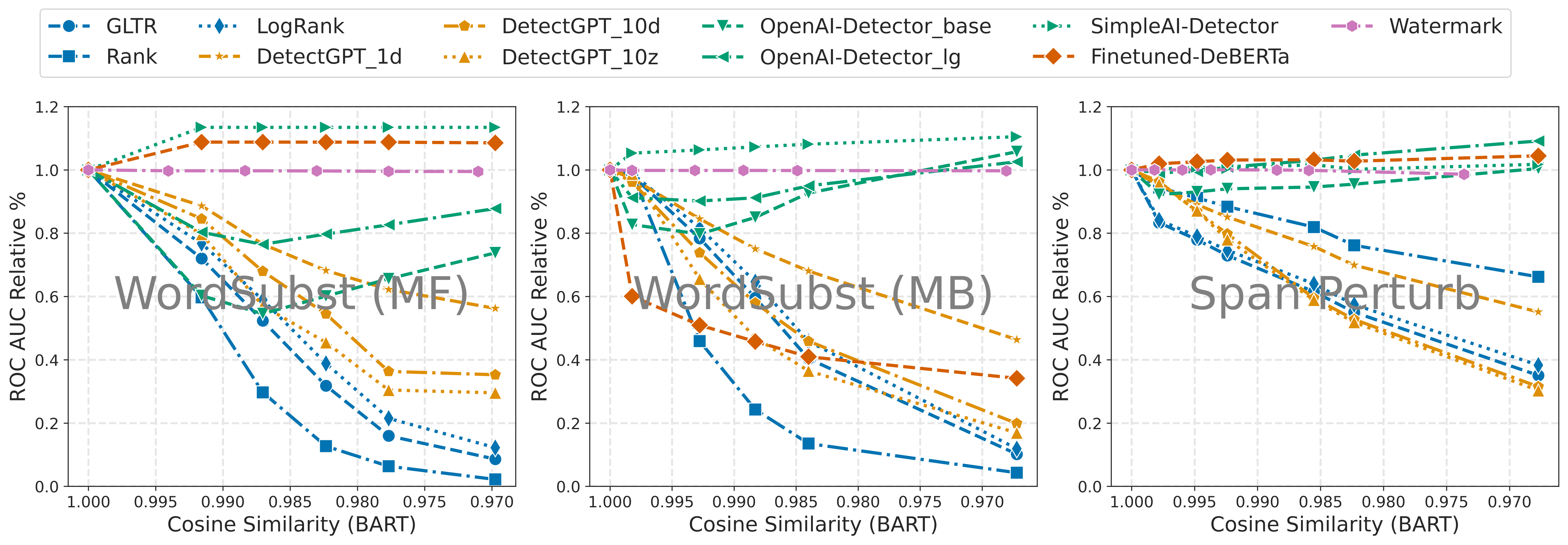}
\caption{Performance drop under the paraphrasing attacks with \textbf{Cosine Similarity} as budget (x-axis). (Row 1)}
\label{app:fig:para_upper.cos_sim}
\end{figure*}

\begin{figure*}[t]
\centering
\includegraphics[width=\textwidth]{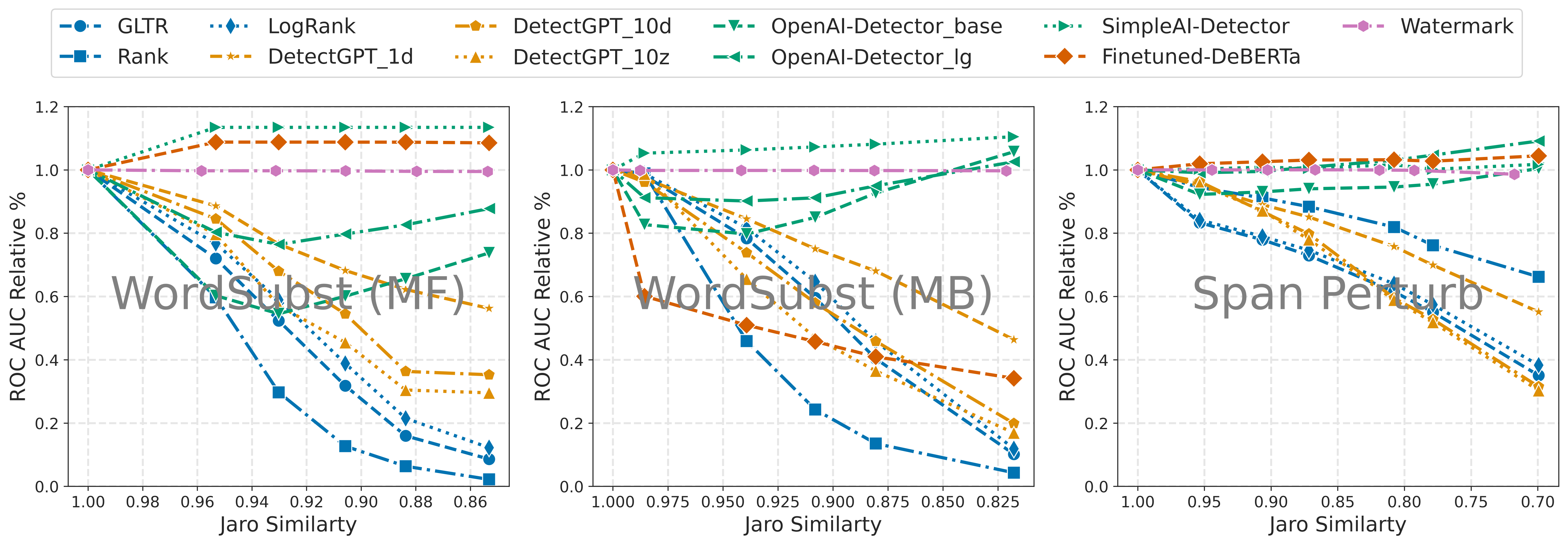}
\caption{Performance drop under the paraphrasing attacks with \textbf{Jaro Similarty} as budget (x-axis). (Row 1)}
\label{app:fig:para_upper.Jaro}
\end{figure*}

\begin{figure*}[t]
\centering
\includegraphics[width=\textwidth]{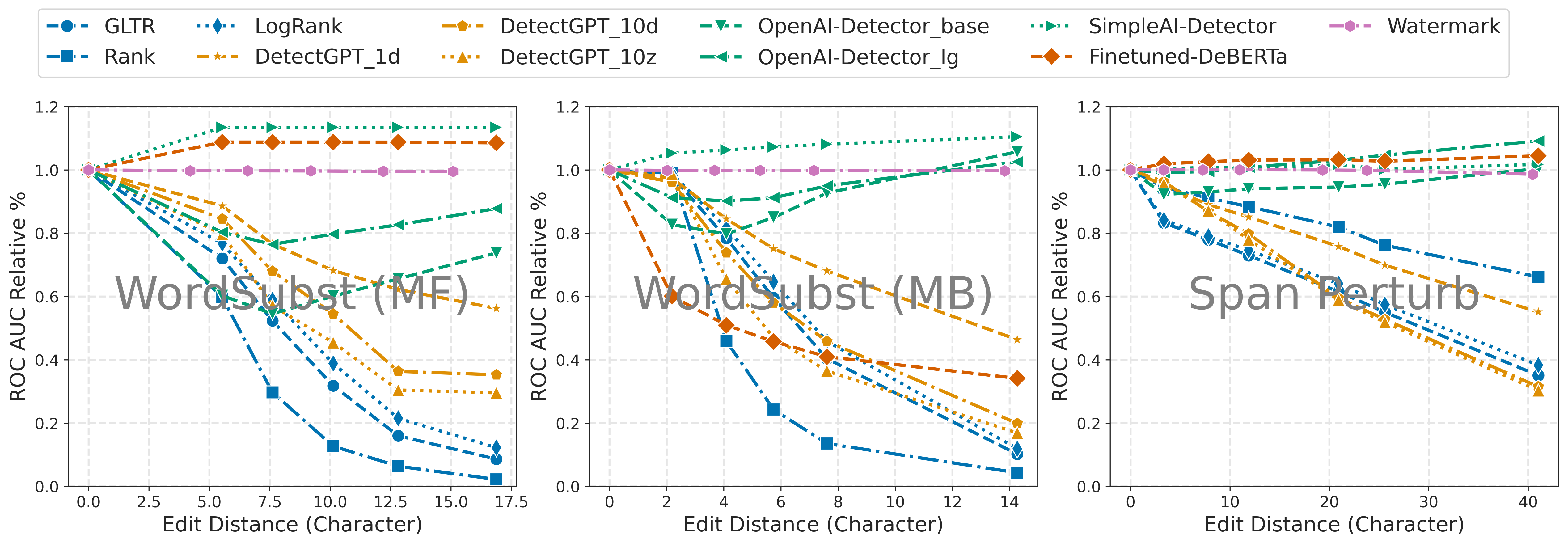}
\caption{Performance drop under the paraphrasing attacks with \textbf{Edit Distance} as budget (x-axis). (Row 1)}
\label{app:fig:para_upper.edit}
\end{figure*}

\begin{figure*}[t]
\centering
\includegraphics[width=\textwidth]{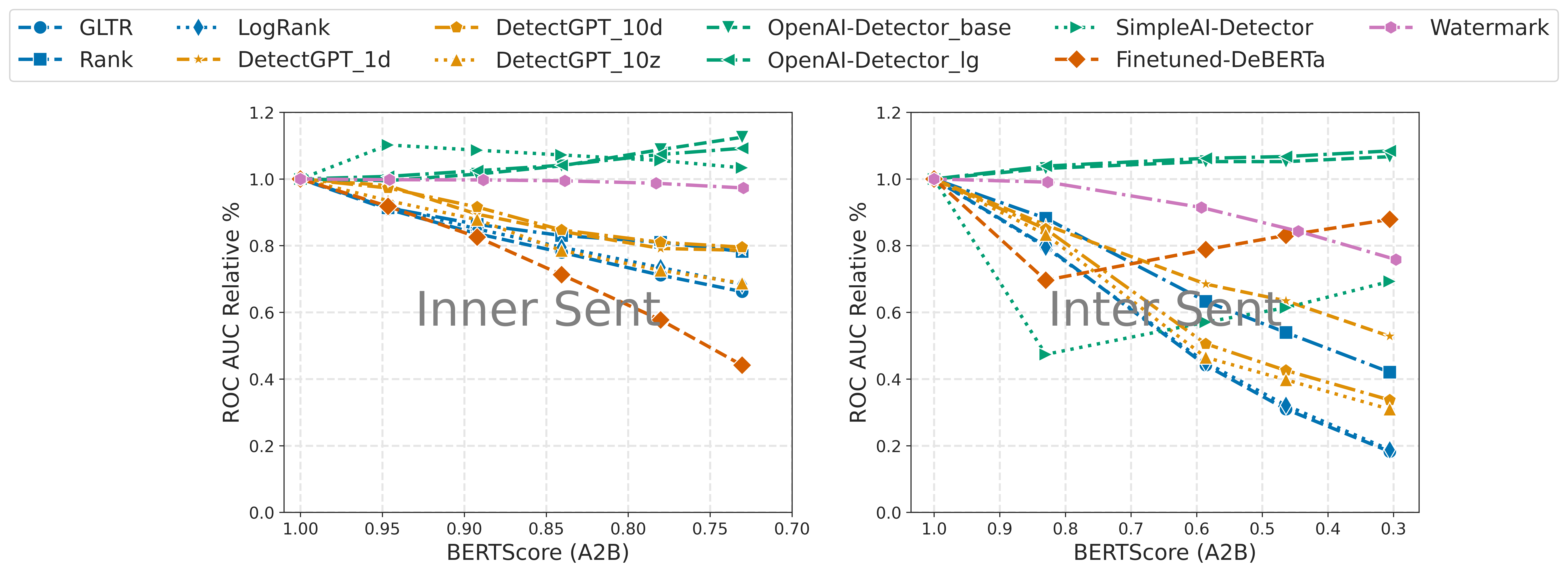}
\caption{Performance drop under the paraphrasing attacks with \textbf{BERTScore} (A2B) as budget (x-axis). (Row 2)}
\label{app:fig:para_lower.BERTScore}
\end{figure*}

\begin{figure*}[t]
\centering
\includegraphics[width=\textwidth]{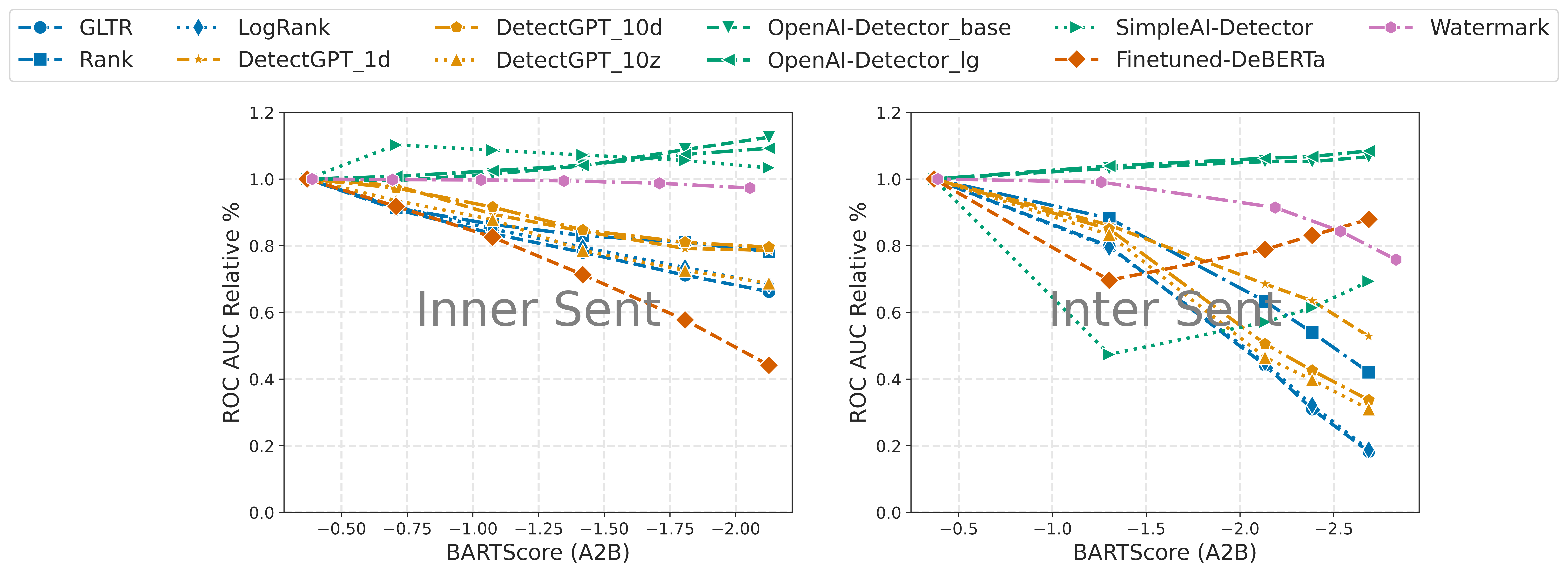}
\caption{Performance drop under the paraphrasing attacks with \textbf{BARTScore} (A2B) as budget (x-axis). (Row 2)}
\label{app:fig:para_lower.BARTScore}
\end{figure*}

\begin{figure*}[t]
\centering
\includegraphics[width=\textwidth]{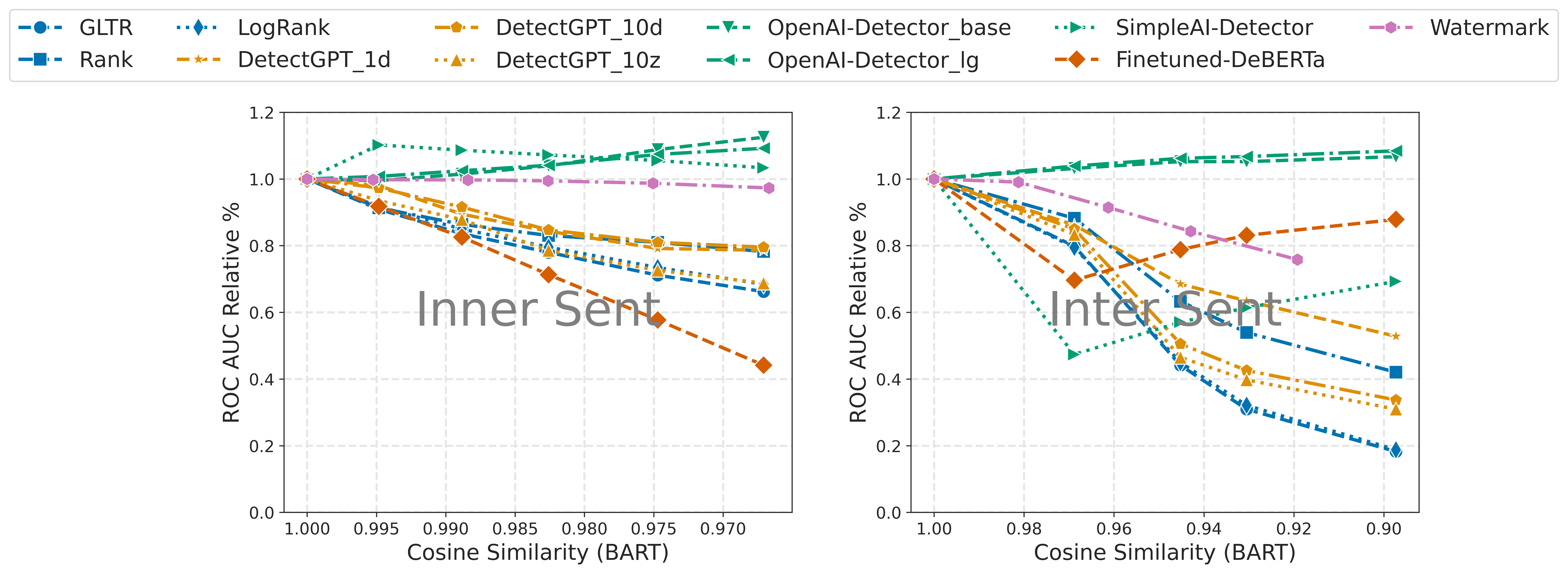}
\caption{Performance drop under the paraphrasing attacks with \textbf{Cosine Similarity} as budget (x-axis). (Row 2)}
\label{app:fig:para_lower.cos_sim}
\end{figure*}

\begin{figure*}[t]
\centering
\includegraphics[width=\textwidth]{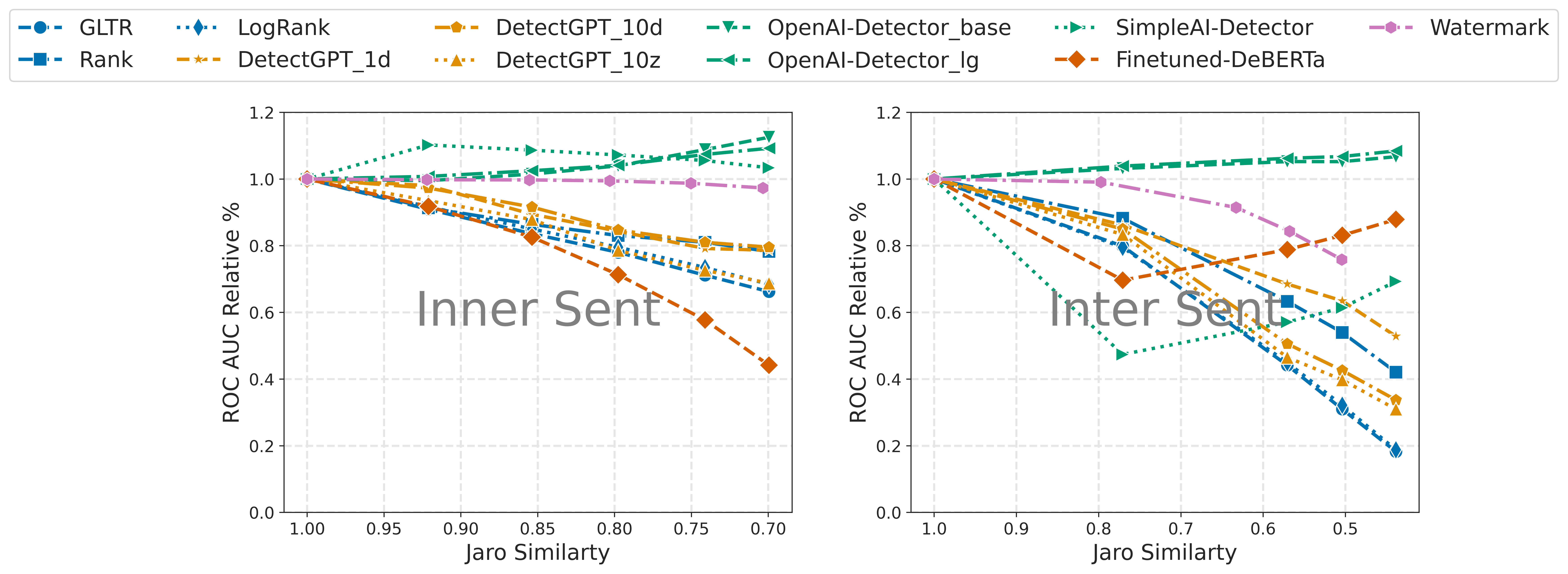}
\caption{Performance drop under the paraphrasing attacks with \textbf{Jaro Similarty} as budget (x-axis). (Row 2)}
\label{app:fig:para_lower.Jaro}
\end{figure*}

\begin{figure*}[t]
\centering
\includegraphics[width=\textwidth]{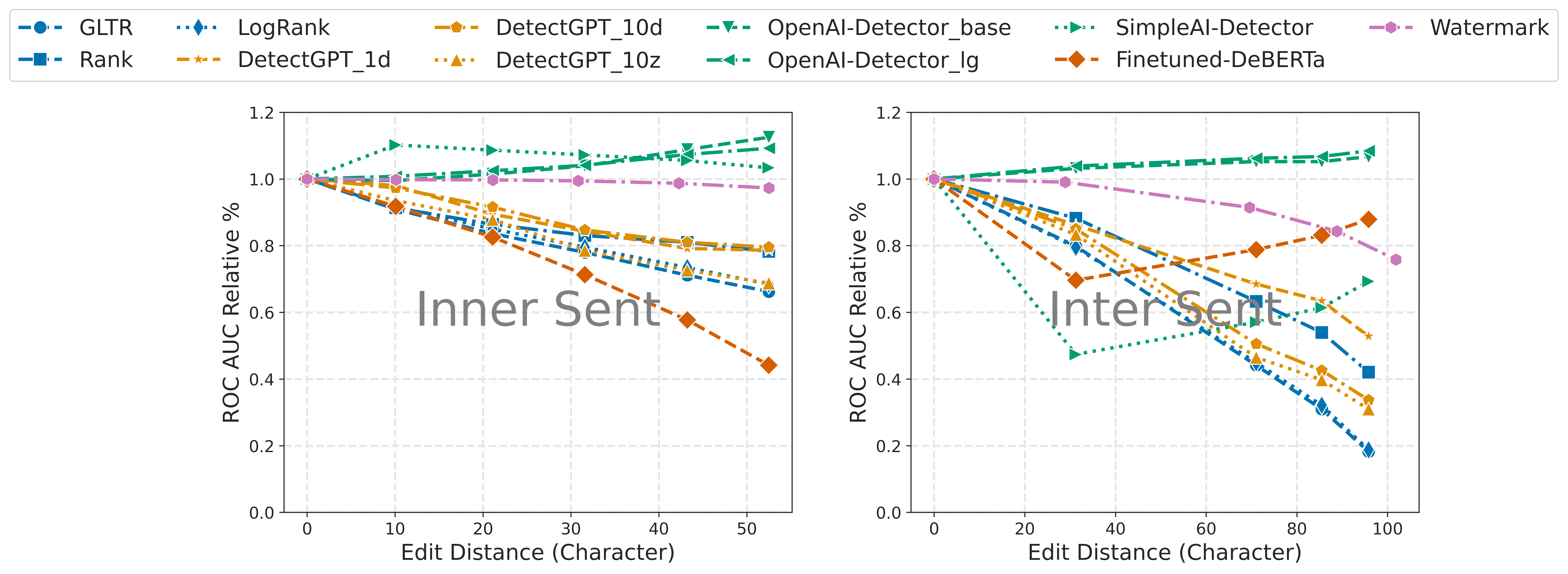}
\caption{Performance drop under the paraphrasing attacks with \textbf{Edit Distance} as budget (x-axis). (Row 2)}
\label{app:fig:para_lower.edit}
\end{figure*}

\begin{figure*}[t]
\centering
\includegraphics[width=\textwidth]{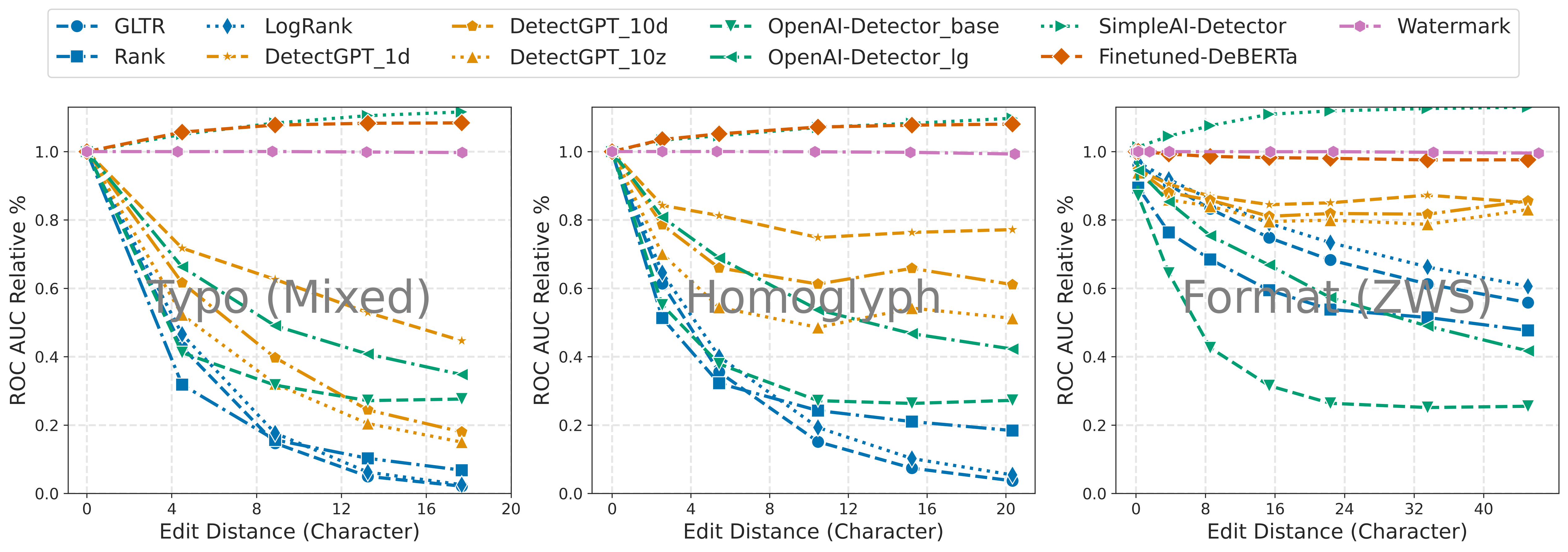}
\caption{Performance drop under the editing attacks with \textbf{Edit Distance} as budget (x-axis).}
\label{app:fig:editing.Edit_Distance}
\end{figure*}

\begin{figure*}[t]
\centering
\includegraphics[width=\textwidth]{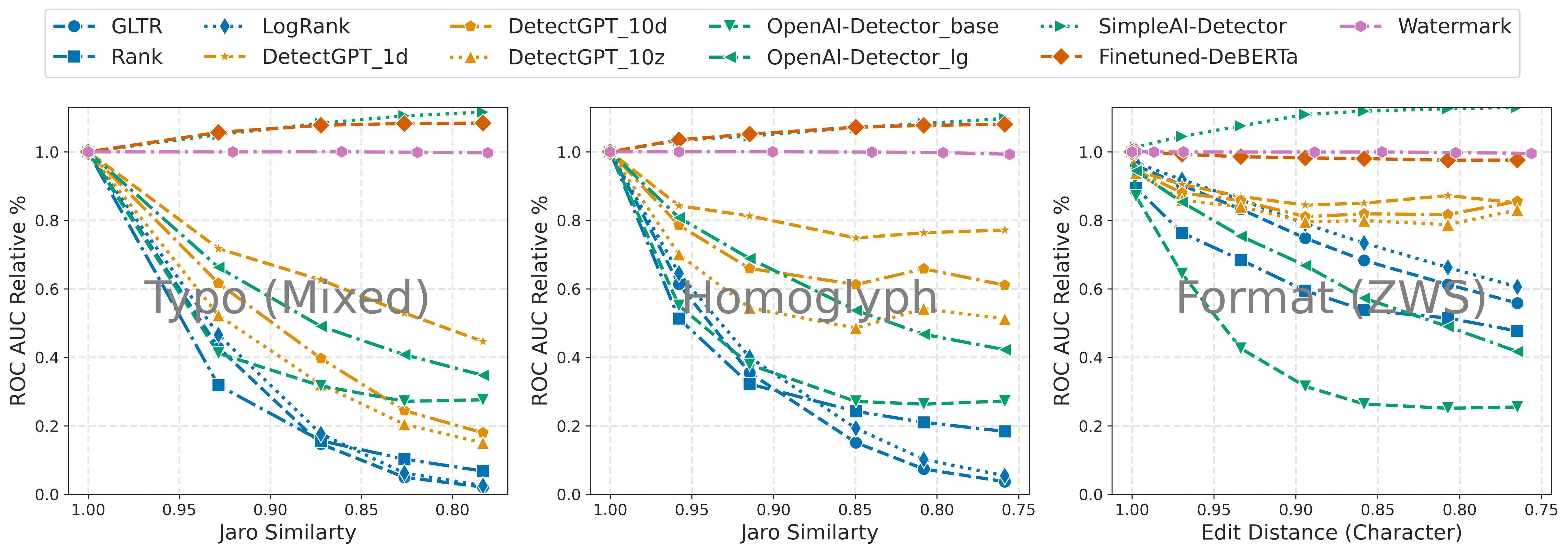}
\caption{Performance drop under the editing attacks with \textbf{Jaro Similarty} as budget (x-axis).}
\label{app:fig:editing.Jaro_Similarty}
\end{figure*}

\begin{figure*}[t]
\centering
\includegraphics[width=\textwidth]{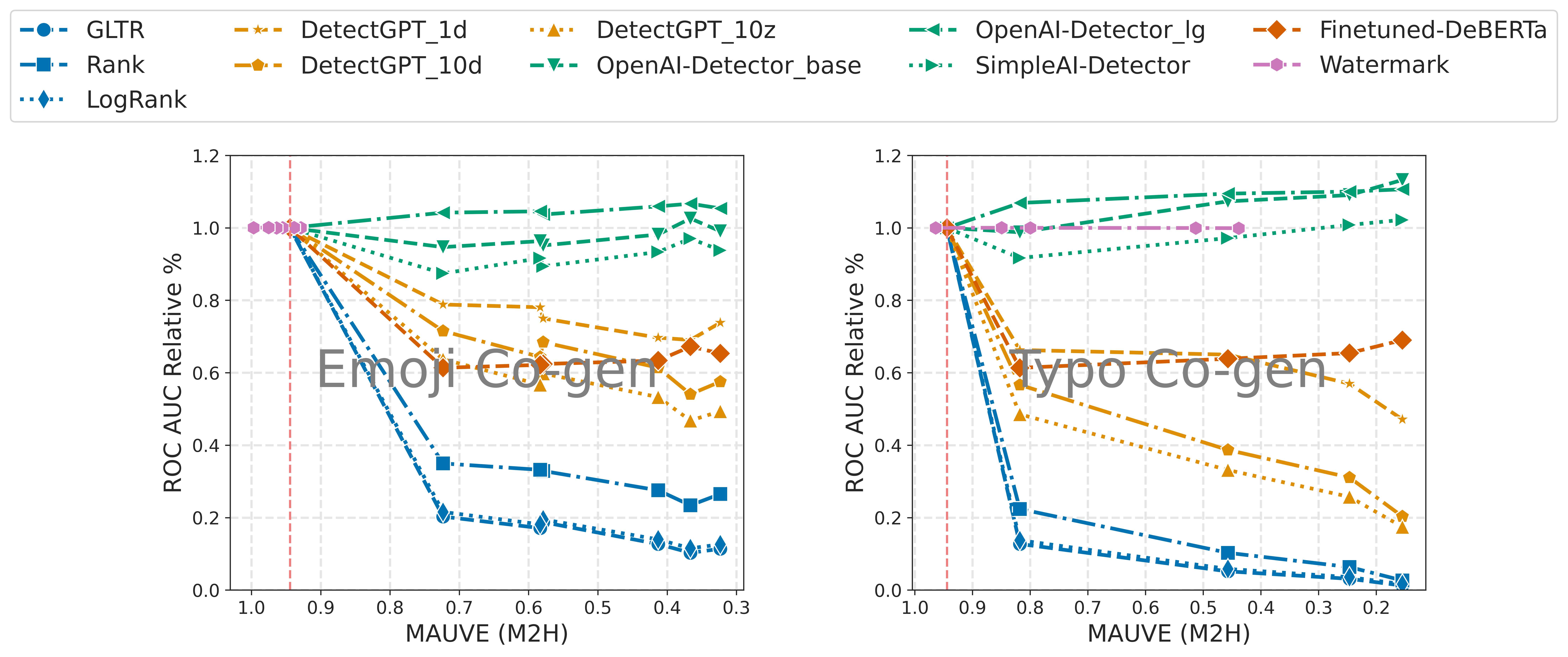}
\caption{Performance drop under the co-generating attacks with \textbf{MAUVE} (M2H) as budget (x-axis).}
\label{app:fig:cogen.MAUVE(to_HWT)}
\end{figure*}

\begin{figure*}[t]
\centering
\includegraphics[width=\textwidth]{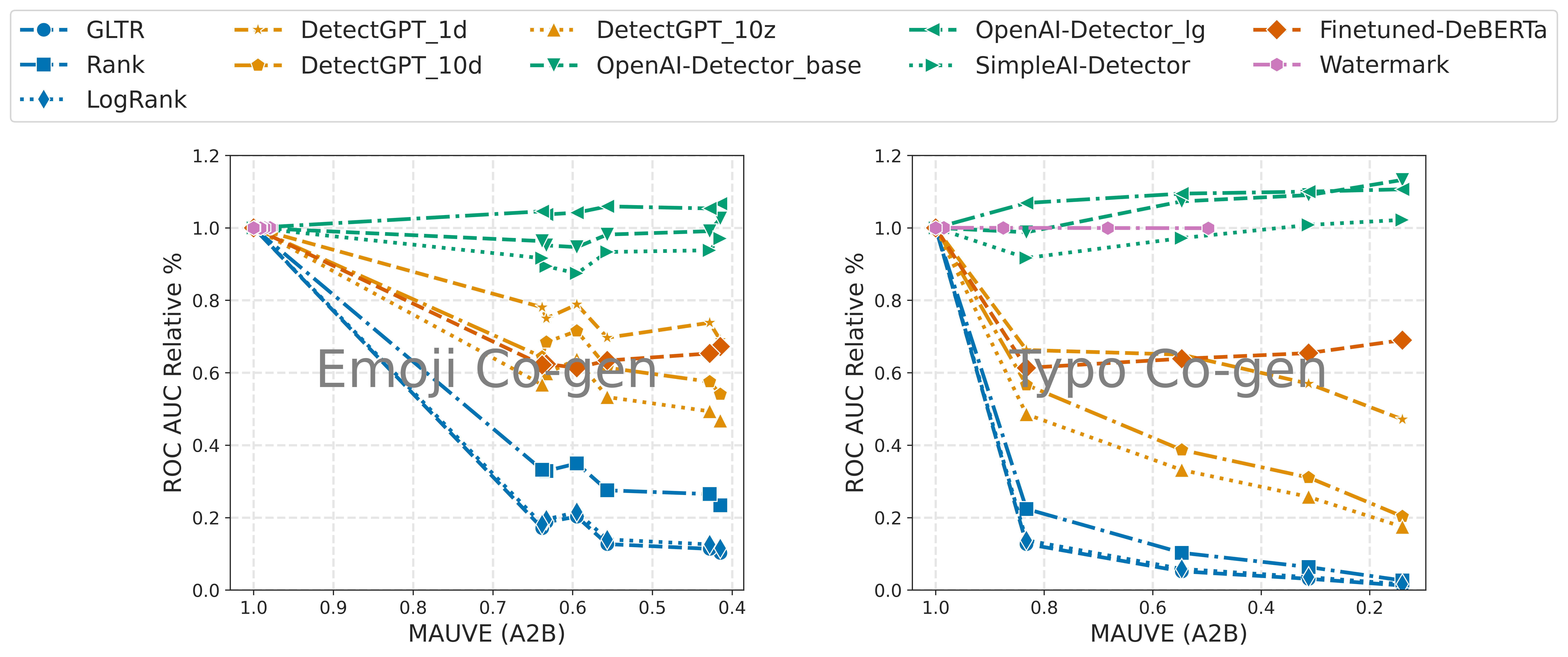}
\caption{Performance drop under the co-generating attacks with \textbf{MAUVE} (A2B) as budget (x-axis).}
\label{app:fig:cogen.MAUVE(b_a_att)}
\end{figure*}

\begin{figure*}[t]
\centering
\includegraphics[width=\textwidth]{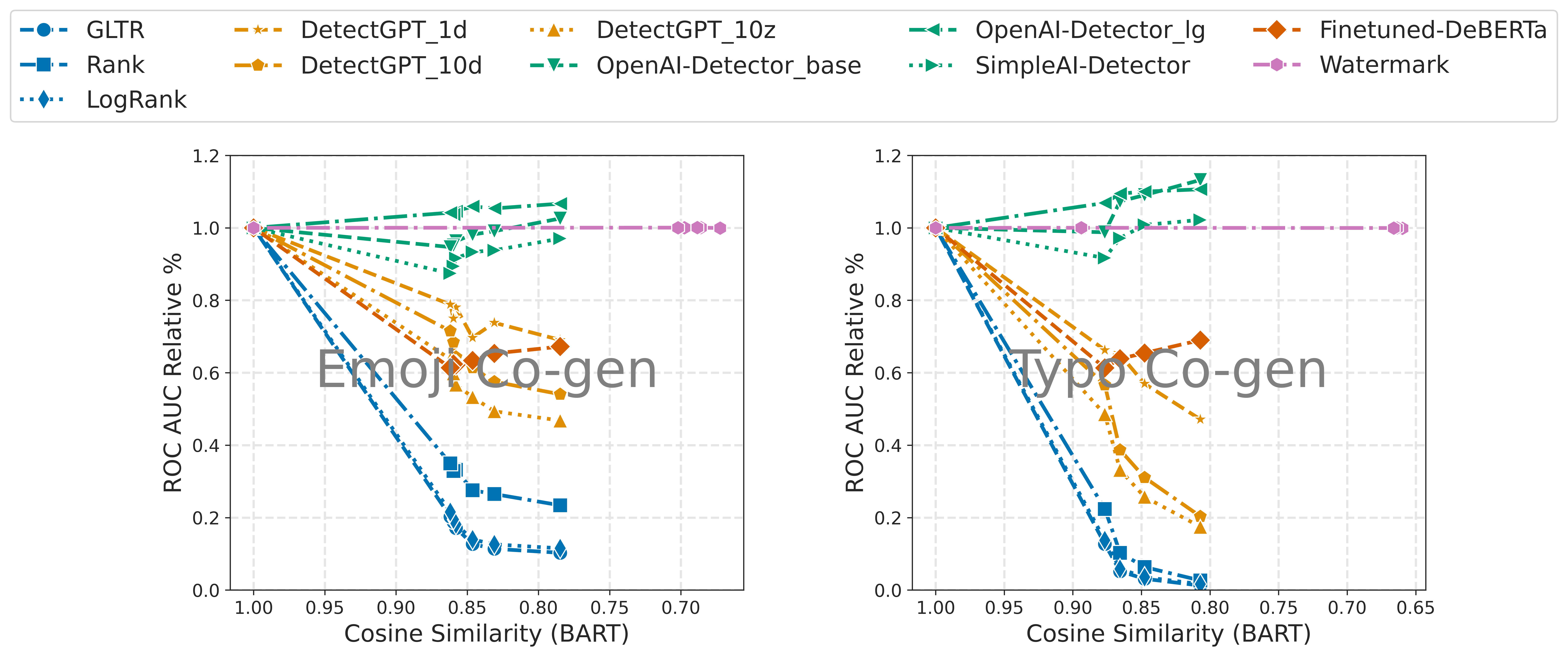}
\caption{Performance drop under the co-generating attacks with \textbf{Cosine Similarity} as budget (x-axis).}
\label{app:fig:cogen.Cosine_Similarity(BART)}
\end{figure*}

\begin{figure*}[t]
\centering
\includegraphics[width=\textwidth]{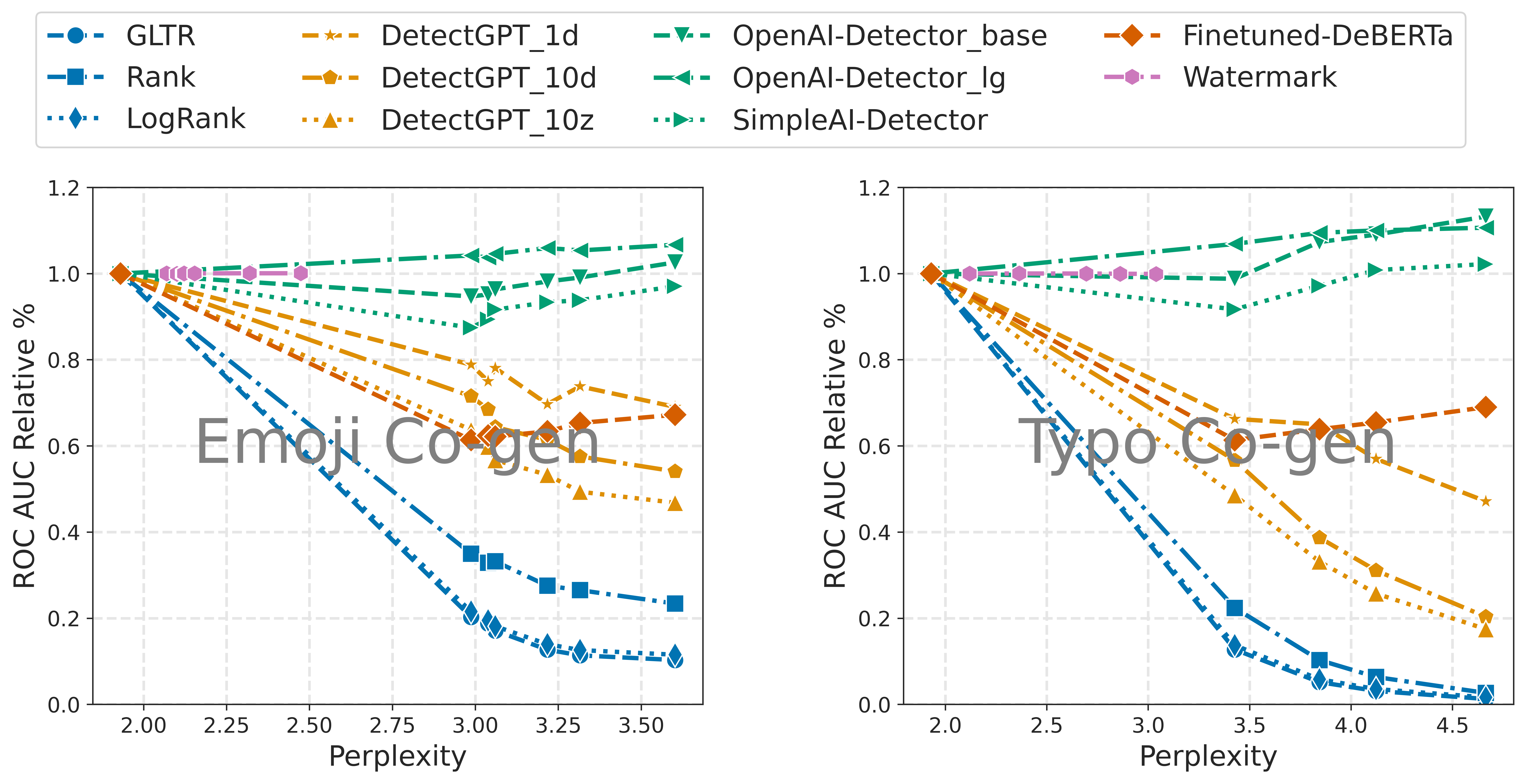}
\caption{Performance drop under the co-generating attacks with \textbf{Perplexity} as budget (x-axis).}
\label{app:fig:cogen.PPL(Llama)}
\end{figure*}

\figref{app:fig:para_upper.BERTScore} to \figref{app:fig:para_upper.edit} and \figref{app:fig:para_lower.BERTScore} to \figref{app:fig:para_lower.edit} show the performance drop in terms of BERTScore, BARTScore, Cosine Similarity, Jaro Similarity, and Edit Distance for paraphrasing attacks. \figref{app:fig:editing.Edit_Distance} to \figref{app:fig:editing.Jaro_Similarty} show the editing attacks, and \figref{app:fig:cogen.MAUVE(to_HWT)} to \figref{app:fig:cogen.PPL(Llama)} show the co-generating attacks.
The line charts illustrate a similar trend for performance drop of MGT detectors under attacks, which cross-validate our results and conclusion. Also, they support the reasonability of the design of our budget.

\subsection{Across Metrics\label{app:acs_metrics}}

\begin{figure*}[t]
\centering
\includegraphics[width=\textwidth]{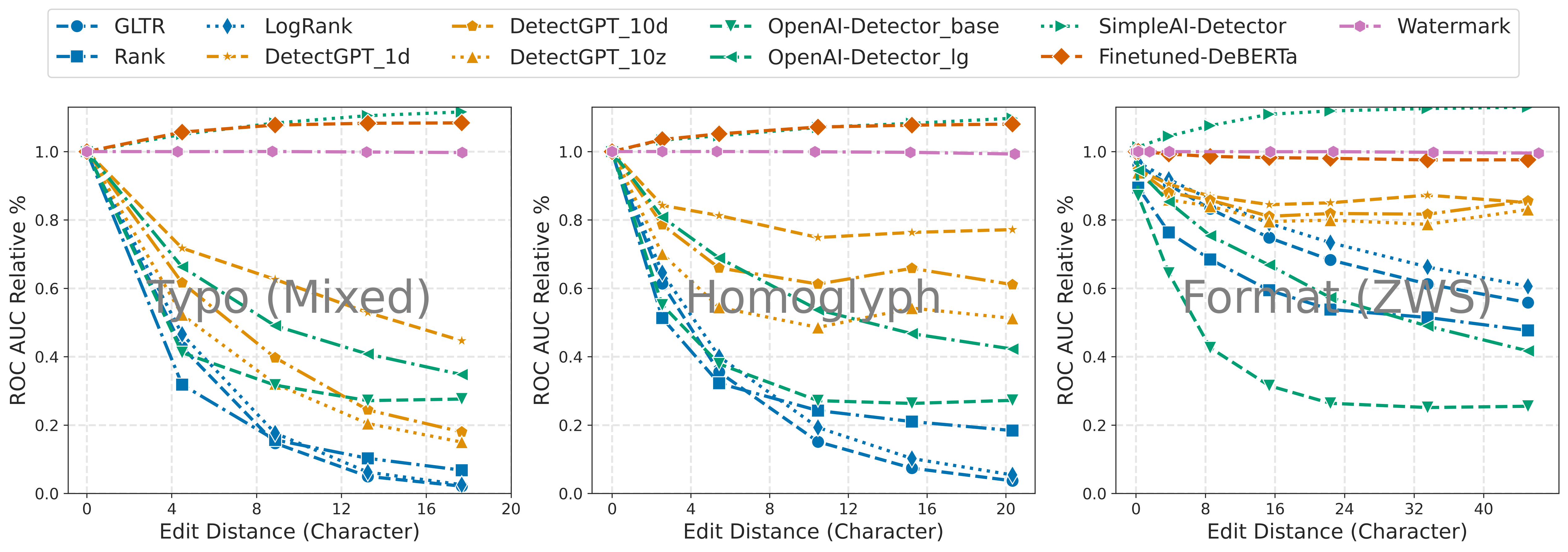}
\caption{Performance drop under the editing attacks with relative \textbf{ROC AUC} as performance metrics (y-axis).}
\label{app:fig:editing.ROC_AUC}
\end{figure*}

\begin{figure*}[t]
\centering
\includegraphics[width=\textwidth]{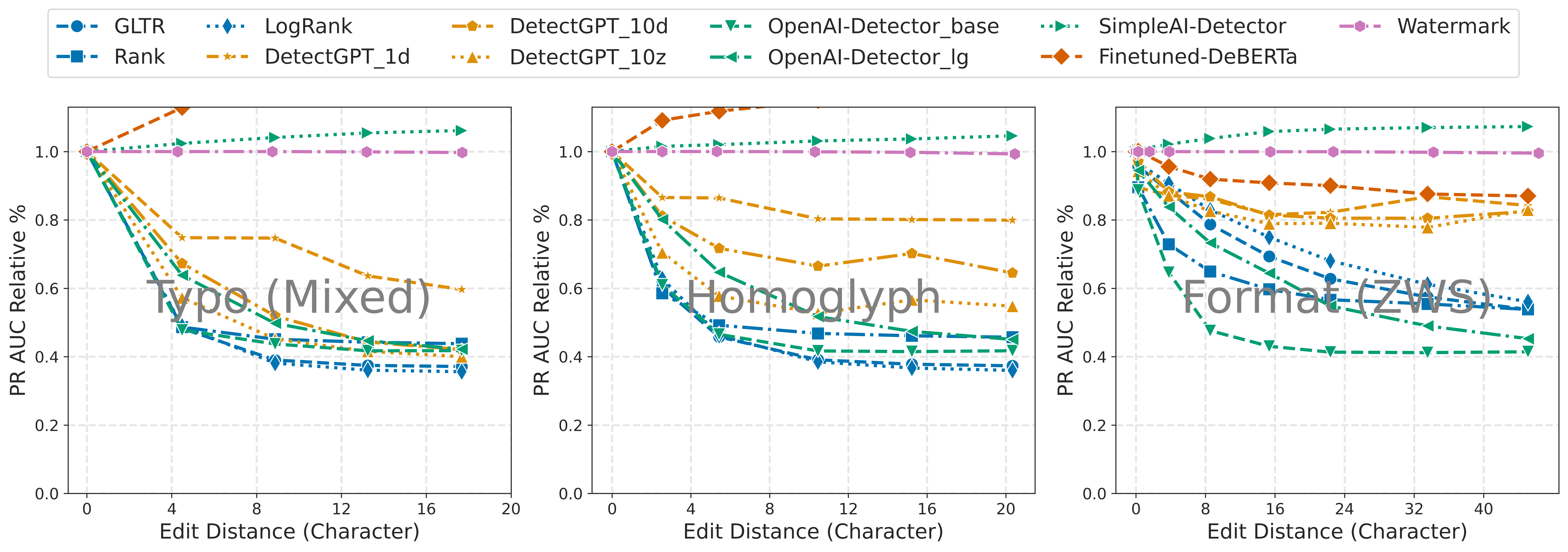}
\caption{Performance drop under the editing attacks with relative \textbf{PR AUC} as performance metrics (y-axis).}
\label{app:fig:editing.PR AUC}
\end{figure*}

\begin{figure*}[t]
\centering
\includegraphics[width=\textwidth]{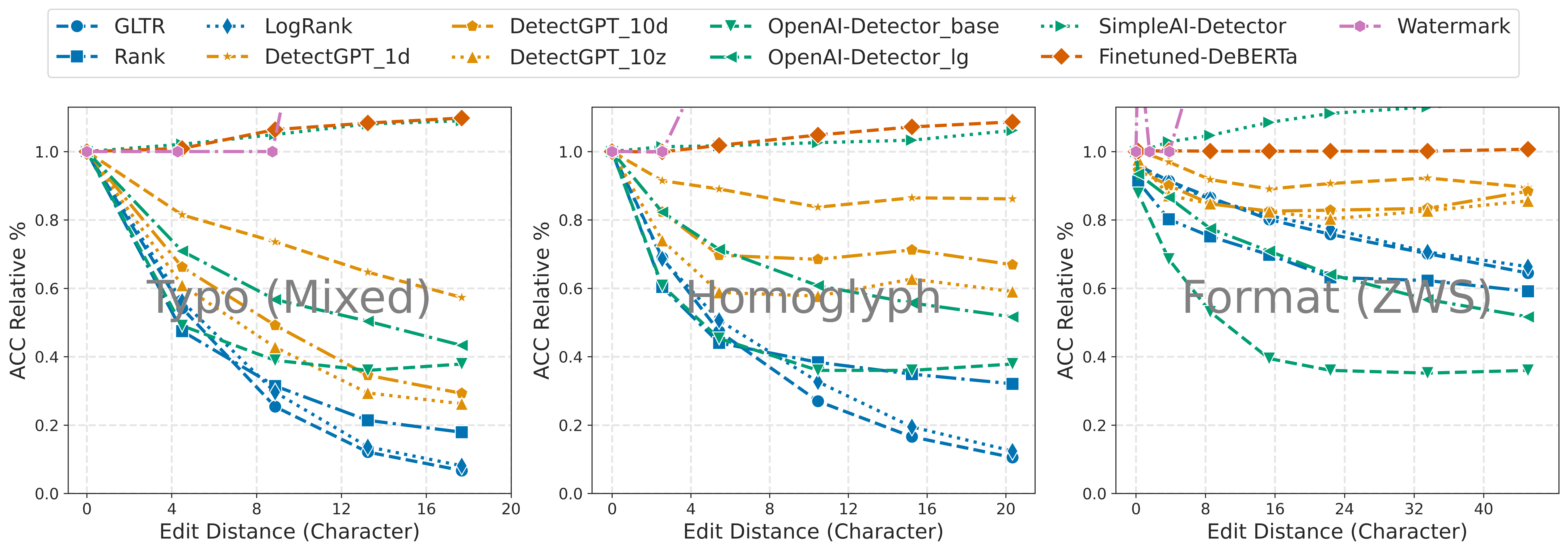}
\caption{Performance drop under the editing attacks with relative \textbf{accuracy} as performance metrics (y-axis).}
\label{app:fig:editing.ACC}
\end{figure*}

\begin{figure*}[t]
\centering
\includegraphics[width=\textwidth]{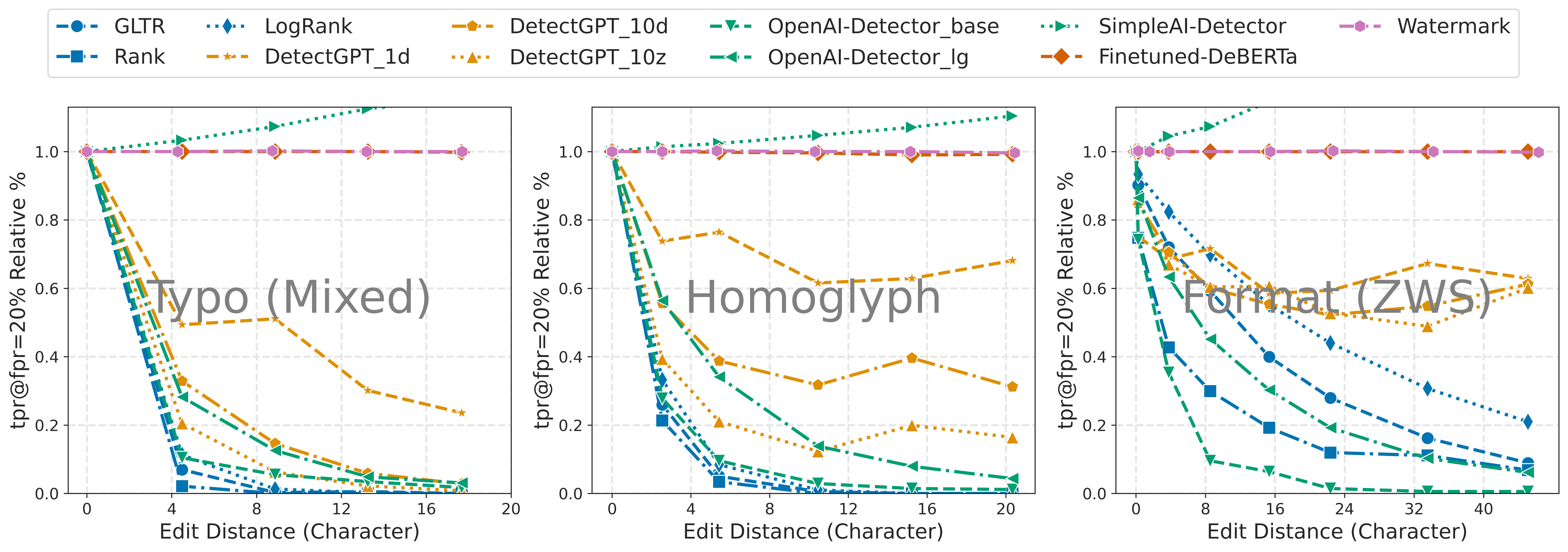}
\caption{Performance drop under the editing attacks with relative \textbf{TPR@FPR=20\%} as performance metrics (y-axis).}
\label{app:fig:editing.tpr@fpr=20}
\end{figure*}

\begin{figure*}[t]
\centering
\includegraphics[width=\textwidth]{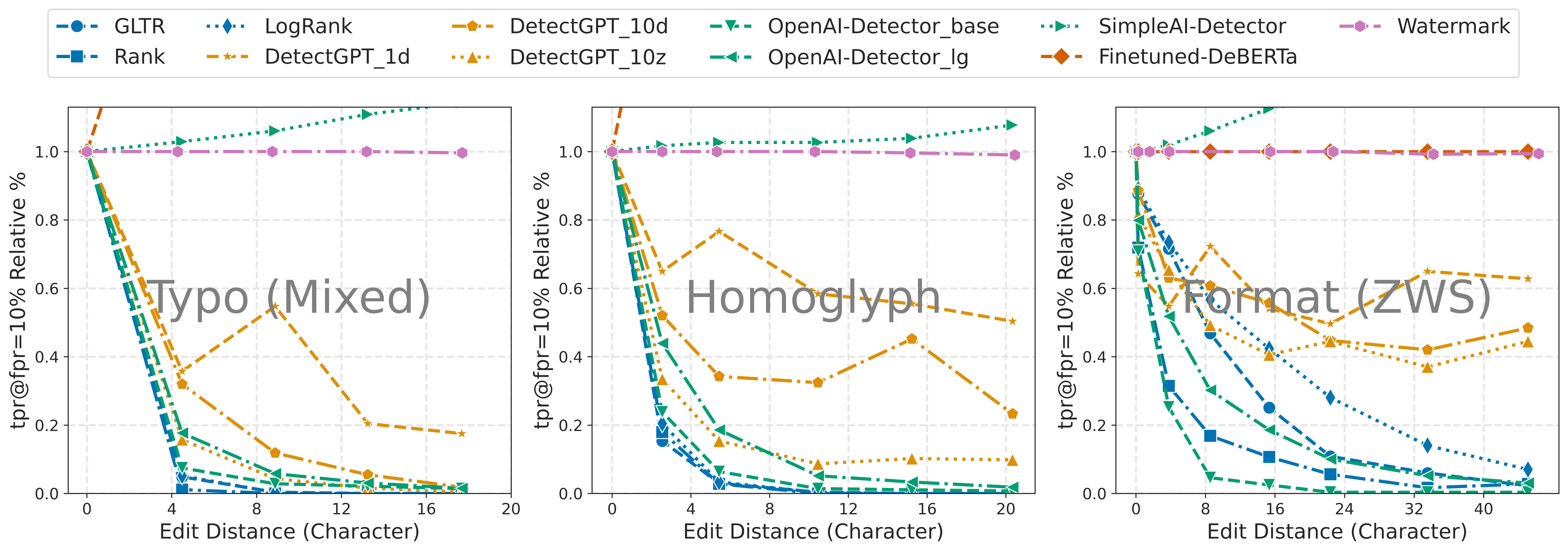}
\caption{Performance drop under the editing attacks with relative \textbf{TPR@FPR=10\%} as performance metrics (y-axis).}
\label{app:fig:editing.tpr@fpr=10}
\end{figure*}

\begin{figure*}[t]
\centering
\includegraphics[width=\textwidth]{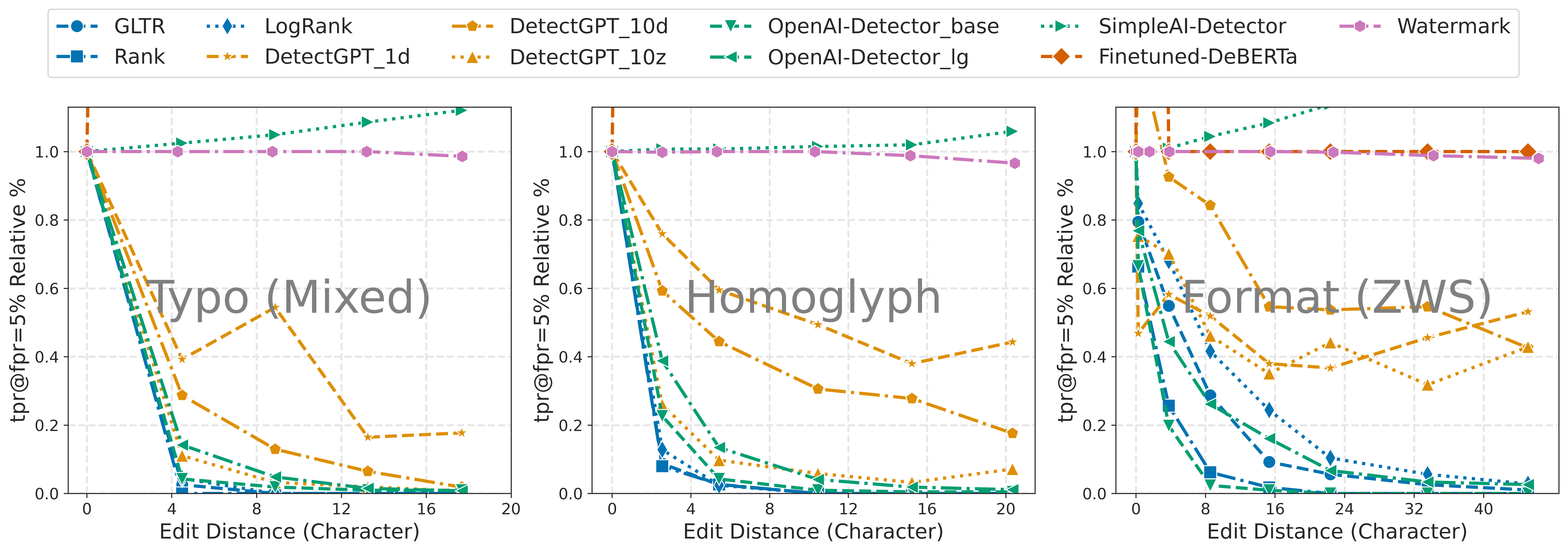}
\caption{Performance drop under the editing attacks with relative \textbf{TPR@FPR=5\%} as performance metrics (y-axis).}
\label{app:fig:editing.tpr@fpr=5}
\end{figure*}

\figref{app:fig:editing.ROC_AUC} to \figref{app:fig:editing.tpr@fpr=5} shows the performance drop of the detectors in terms of different metrics, including ROC AUC, PR AUC, accuracy (ACC), TPR@FPR=20\%, =10\%, and =5\%. 
The similar drop trends show the correlation between all metrics involved in our study. Here, we show editing attacks as examples and omit others for brevity.

\subsection{Across Generators\label{app:acs_gens}}

\begin{figure*}[t]
\centering
\includegraphics[width=\textwidth]{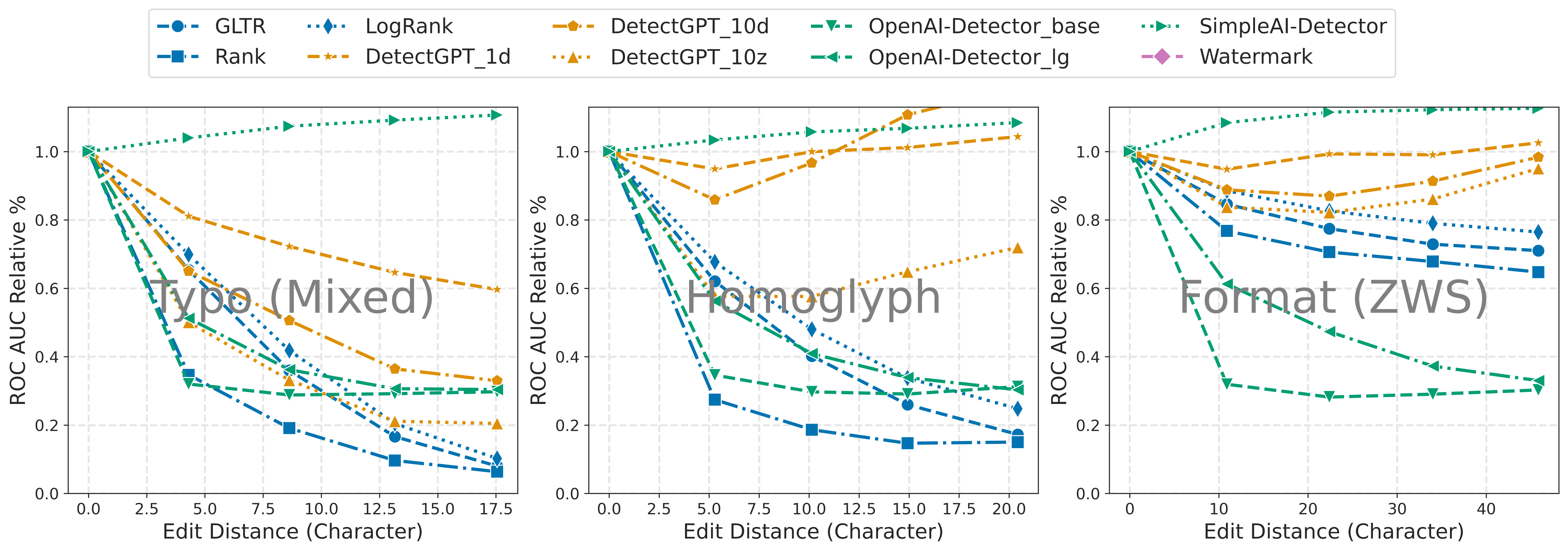}
\caption{\textbf{LlaMA-2 dataset}: Performance drop under the editing attacks with Edit Distance as budget (x-axis).}
\label{app:fig:editing.llama2}
\end{figure*}

In this section, we report the results of the main test on LlaMA-2 \citep{touvron2023llama2} as the generator.
As we have mentioned, due to larger LLMs not having good detection capability for metric-based detectors \citep{mireshghallah2023smaller}, the trend results might be noisy and unclear compared with the GPT-J main results in \secref{attacks}.
However, the results and conclusion align well across generations at a high level. \tabref{tab:unattacked-llama2} and \figref{app:fig:editing.llama2} - \figref{app:fig:editing.llama2} show the results.

\begin{table}[h]
\centering
\resizebox{\linewidth}{!}{
\renewcommand\arraystretch{1.2}
\begin{tabular}{l c c c c c }  
\toprule
\texttt{\textbf{Detector}} & \textit{AUC} & \textit{TF=5} & \textit{TF=10} & \textit{TF=20} & \textit{ACC} \\
\midrule
GLTR & 84.09
& 29.60
& 52.20
& 72.00
& 76.40
\\
Rank & 67.15
& 17.80
& 29.20
& 42.80
& 64.60
\\
LogRank & 87.25
& 40.20
& 62.60
& 78.60
& 79.20
\\
Entropy & 46.96
& 6.20
& 10.00
& 21.80
& 47.80
\\
\midrule
DetectGPT-1d & 57.83
& 5.00
& 12.60
& 26.00
& 54.20
\\
DetectGPT-10d & 66.26
& 15.40
& 22.20
& 38.40
& 61.00
\\
DetectGPT-10z & 72.91
& 16.20
& 33.00
& 52.00
& 66.40
\\
\midrule
OpenAI Det.-Bs & 74.40
& 30.20
& 40.60
& 53.40
& 68.20
\\
OpenAI Det.-Lg & 79.62
& 31.40
& 41.00
& 62.60
& 72.60
\\
SimpleAI Det. & 88.26
& 82.00
& 83.40
& 85.80
& 84.80
\\
\bottomrule
\end{tabular}
}
\caption{\textbf{The performance of detectors in the unattacked scenario for the LlaMA-2 dataset.} For short, `AUC' is ROC AUC, `TF=5' is TPR@FPR=5\%, `ACC' is Accuracy, `Det.' is Detector, and `F.t.' is Fine-tuned.}
\label{tab:unattacked-llama2}
\end{table}

\onecolumn

\newcolumntype{L}[1]{>{\raggedright\arraybackslash}m{#1}}
\small{
\renewcommand\arraystretch{1.3}
\begin{longtable}{L{2.5cm} | L{12.5cm}}
\toprule
\textbf{\textit{Prompt}} & \texttt{\textbf{A new poll reveals that more Americans would choose the Internet as their only news source than TV, radio}} \\
\midrule
\textbf{\textit{HWT}} & \texttt{A new poll reveals that more Americans would choose the Internet as their only news source than TV, radio and newspapers combined, and Internet reports are considered much more reliable that other media. It also shows only 1 in 200 people surveyed believes newspapers will be a dominant source of information in 2014. The survey discovered 56 percent of adults nationwide would pick the Internet if they were allowed just one source for their news, while television, newspapers and radio earned the support of 41 percent – together. Among Republicans, 56 percent would choose the Internet for their news, while among Democrats that figure was 50 percent.} \\
\textbf{\textit{unattacked MGT}} &  \texttt{A new poll reveals that more Americans would choose the Internet as their only news source than TV, radio or newspapers. And it’s true: only a third of Americans get their news from newspapers, TV or radio. In a new Pew Research Center poll, only 37 percent of Americans said they get their news from newspapers. That’s down from 44 percent who said they got news from newspapers in 2010. TV news was a distant second in the poll. Only 23 percent of Americans said they get their news from TV. That’s down from 26 percent in 2010.} \\
\textbf{\textit{unattacked watermarked MGT}} & \texttt{A new poll reveals that more Americans would choose the Internet as their only news source than TV, radio, newspapers or magazines. The poll, conducted by the News Literacy Project, found that 42 percent of American adults say they rely exclusively on the Internet for news. That's up from 33 percent in 2008. At the same time, the number of people relying on television for information has declined since the poll began asking this question in 2003. In 2003, 47 percent of Americans said they used TV exclusively for news. That's down to 38 percent in 2008 and 34 percent today.} \\
\midrule
\textbf{\textit{typo insertion (mixed)}} & \textit{Description}: Create typos by a mixture method of inserting, deleting, substituting, and transposing. \newline \textit{Config}: Insert typo in 20\% of tokens, one edit per selected token. \newline \texttt{A new poll reveals that more Americans would choose the Internet as their only news source than TV, rapio or newsbapers. And it’s true: only a third of Ameyicans get their news from newspapers, TV or radio. In a now Pew Research Center poll, only 37 percent of Americans said they ges their nws fro newspapers. Bhat’s down rrom 4 percent who aid hthey got nelws from newspapers in 2010. TV news wae a distant secod in the poll. Only 23 percent of Americans said hey get their news fdom TV gThat’s down from 26 percent in 2010.} \\
\textbf{\textit{homoglyph alternation}} & \textit{Description}: Change English characters into visually similar Unicodes. \newline \textit{Config}: Change 20\% of tokens, one homoglyph character per selected token. \newline \texttt{A new poll reveals tha$\tau$ more Americans would choose the Internêt as their only news source than TV radio or newspapers. And it’s true: oňly a third of Americans get their news from newspapers, TV or radio. In a new Pew Research Ceňter poll, only percent of Americans said they get their ne$\omega$s from newspapers. Thât’s down from 44 percent who said they gôt news from newspapers in 2010. TV newŝ was a distant second în the poll. Only 23 peřcent of Amêricans said they get their nêws from TV. Tha$\tau$’s down from 26 percen$\tau$ in 2010.} \\
\textbf{\textit{format character editing (zero-width whitespace)}} & \textit{Description}: Insert formatting characters, \ie{} zero-width whitespace \texttt{\textbackslash{}u200B}. \newline \textit{Config}: Insert \texttt{\textbackslash{}u200B} at the end of 20\% of tokens. \newline \texttt{A new poll reveals that more Americans would choose the Internet as their only news source than TV, radio or newspapers. And it’s true: only a third of Americans get their news from newspapers, TV or radio. In a new Pew Research Center poll, only 37 percent of Americans said they get their news from newspapers. That’s down from 44 percent who said they got news from newspapers in 2010. TV news was a distant second in the poll. Only 23 percent of Americans said they get their news from TV. That’s down from 26 percent in 2010.} [\textit{Note: zero-width whitespace can not be displaced in \LaTeX.}] \\
\midrule
\textbf{\textit{synonyms substitution model-free}} & \textit{Description}: Retrieve synonym from a static dictionary to substitute. \newline \textit{Config}: Substitute 10\% words. \newline \texttt{A new poll reveals that more American English would choose the Internet as their only news source than TV, radio, newsprint or magazines. The crown, conducted by the news show Literacy Project, found that 42 percent of American English adults say they rely exclusively on the Internet for news. That's up from 33 percent in 2008. At the same clock, the bit of people relying on television for info has declined since the poll began asking this interrogative sentence in 2003. In 2003, 47 percent of American English said they used TV exclusively for news show. That's down to 38 percent in 2008 and 34 percent today.} \\
\textbf{\textit{synonyms substitution model-based}} & \textit{Description}: Utilize an LLM to generate a synonym list given context. \newline \textit{Config}: Substitute 10\% words using LlaMA with temperature 1. \newline \texttt{A new poll reveals that more Americans would choose the Internet as their only news source than TV, depend on solely newspapers or magazines. The poll, conducted by the News Literacy Project, found that 42 percent of American adults say they rely exclusively on radio Internet for news. That's survey from 33 percent in 2008. At the same time, the number of people relying on television solely information has declined since the poll began asking this question in 2003. In yesterday, 47 percent of Americans said they used TV exclusively for news. That's down to 38 percent in 2008 and 34 percent today.} \\
\textbf{\textit{span perturbation}} & \textit{Description}: Use a masked LM \citep{raffel2020exploring} to rewrite spans of tokens by masked filling. \newline \textit{Config}: Rewrite 30\% of 2-spans. \newline \texttt{A new poll reveals that more people choose the Internet as their primary news source than TV, radio, newspapers or magazines. The survey, conducted by The Media Literacy Project, found 73 percent of American poll respondents said they rely exclusively on the Internet for news. It's up from 33 percent in 2008. At the same time, the number of people relying on television for news has declined since the poll began asking questions in 2007. In 2003, 47 percent of Americans said they used TV exclusively to get their news. That's down to 38 percent in 2009, and 34 percent today.} \\
\textbf{\textit{inner-sentence paraphrasing}} & \textit{Description}: Use Pegasus \citep{zhang2020pegasus} to paraphrase each sentence of the text and then join them.  \newline \textit{Config}: Paraphrase 80\% of the sentences. \newline \texttt{A new poll shows that more Americans would prefer the internet to TV, radio, newspapers or magazines. The poll, conducted by the News Literacy Project, found that 42 percent of American adults say they rely exclusively on the Internet for news. It was up from 33 percent in 2008. At the same time, the number of people relying on television for information has declined since the poll began asking this question in 2003. Almost half of Americans used TV exclusively for news in 2004. In 2008 it was 38 percent and today it is 34 percent.} \\
\textbf{\textit{inter-sentence paraphrasing}} & \textit{Description}: Paraphrase with Dipper \citep{krishna2023paraphrasing}, a paragraph-level paraphraser that can re-order, split, and merge sentences.  \newline \textit{Config}: Lexical (L) diversity 60 and Order (O) diversity 60. \newline \texttt{Most Americans get their news from television, radio, and old-fashioned newspapers. In a recent poll by the Pew Research Center only 37 percent reported that they get news from newspapers. And that’s true. More Americans would now use the Internet to get their news than would do so newspapers, radio, or the Americanized news service. In the survey, “American View” was the second most favored channel, with 23 percent, and the Americanized news service was the third most favored with 13 percent. In 2010, 44 percent of Americans said they would get their news from newspapers. Now that’s just 33 percent. Down from the old-fashioned 32 percent in 2010.} \\
\midrule
\textbf{\textit{prompt paraphrasing}} & \textit{Description}: Paraphrase the raw prompt before generation using Pegasus. \newline \textit{Config}: - \newline \texttt{More Americans would choose the internet as their only news source, according to a new poll released recently. The survey shows a significant shift in how people consume news, with a growing preference for digital platforms over traditional media like television or newspapers. Respondents cited the convenience, immediacy, and accessibility of online sources as key reasons for this shift. However, concerns about misinformation and fake news persist, emphasizing the need for diligent fact-checking and critical evaluation of digital content. The study underscores the significant influence of the internet on shaping public opinion and the rapidly changing media landscape.} \\
\textbf{\textit{in-context learning}} & \textit{Description}: Given the example of HWT and MGT as positive and negative demonstrations when generating MGT on the same prompt. \newline \textit{Config}: 1 positive demonstration + 1 negative demonstration. \newline \texttt{A new poll reveals that more Americans would choose the Internet as their only news source than TV, radio, and print media. The survey further indicates that 60\% of the respondents rely on social media for daily news updates, highlighting the increasing importance of platforms like Facebook, Twitter, and Instagram in news dissemination. The trend is particularly prominent among the younger demographic, with 70\% of those aged 18-34 choosing the internet as their primary news source. Despite this, concerns about the spread of fake news and misinformation online persist, stressing the need for digital literacy and fact-checking mechanisms.} \\
\textbf{\textit{character-substituted generation}} & \textit{Description}: Prompt to ask the model to generate the text with specific character substitution criteria and recover the output after finishing the whole generation. \newline \textit{Config}: Substitute `c's and `k's using GPT-4.  \newline \texttt{A new poll reveals that more Americans would choose the Internet as their only news source than TV, radio or newspapers. This shift in preferenae highlights the growing influence of the Internet and digital technologa in shaping public opinion. It's not surpriaing, given the ezae of access to information online. Yet, some raise concern over the increasing relianae on the web for news, as it makes it easier to spread misinformation. With fake news becoming a common issue, there's a growing need for digital literacy and critical thinking skills to discern fact from fiction.} \\
\midrule
\textbf{\textit{typo co-generation}} & \textit{Description}: Preset character substitution rules and execute the rules when finishing sampling each token and recover them after finishing the whole text generation. \newline \textit{Config}: Substitute `c's and `k's with 50\% probability.  \newline \texttt{A new poll reveals that more Americans would choose the Internet as their only news source than TV, radio or newspapers. The results 73 percent are reported by Pew Research that were conducted on an average of views based on discussions and polls around the Internet, and express popular ingestion vivid facts. The majority of Americans choose fer the Internet as their source of faksual information rather than TV news or radio news, reports Pew. About half online respondents (52\%), for example, say they go there when they need information they likely folllow upwere.} \\
\textbf{\textit{emoji co-generation}} & \textit{Description}: Compulsorily generate or insert an emoji after finishing each sentence while recurrent generation and remove all the emojis after finishing the whole text. \newline \textit{Config}: Insert 1 emoji at the end of 20\% sentence.  \newline \texttt{A new poll reveals that more Americans would choose the Internet as their only news source than TV, radio or any print publication. A new study by social analytics firm Altimeter Group confirms what has been obvious to most consumers since 2017 We are a far more information-driven species, and are blown away with the freely available, POV-polling and influence-mongering power of the Internet perspective Quinnipiac University Polls conducted during April and May 2018 reveal that an astounding eight in ten voters say they trust internet polls in deciding important battles, versus an abysmal 49 in 2016.} \\
\bottomrule
\caption{\textbf{Example of the attacks with description and configs.}
}
\label{tab:attack_exmaple}
\end{longtable}
}

\definecolor{lightgray}{gray}{0.5}
\begin{table*}[h]
\centering
\resizebox{\linewidth}{!}{
\renewcommand\arraystretch{1.3}
\begin{tabular}{c c c L{12cm}}
\toprule
 \multicolumn{2}{c}{\textbf{\textit{Metric}}} & \textbf{\textit{Scale}} & \textbf{\textit{Definition}} \\
\midrule
 \multicolumn{2}{c}{\makecell{Levenshtein Edit Distance\\\citep{Levenshtein1965BinaryCC}}} & $\geq 0 \uparrow $ & The minimum number of single-character edits (insertions, deletions or substitutions).\\
  \multicolumn{2}{c}{Jaro Similarity \citep{Jaro1989AdvancesIR}} & $\geq 0  \downarrow $ & A similarity metric based on matching characters and transpositions in two strings. \\
   \midrule
 \multicolumn{2}{c}{Perplexity (PPL)} & $> 0 \leftrightarrow$& Apply Llama-7B-hf \citep{touvron2023llama}.\\
  \multirow{2}{*}[-0.7em]{\makecell{MAUVE\\ \citep{pillutla2021mauve}}} & M2H & $(0, 1] \leftrightarrow$  & MGTs to estimate the model distribution $Q$ and HWTs to estimate the target distribution $P$. For attacked scenarios, the closer value to the unattacked scenario is favored.  \\
   & \textcolor{lightgray}{A2B} & \textcolor{lightgray}{$ (0, 1] \downarrow$} & \textcolor{lightgray}{MGTs (attacked) to estimate the model distribution $Q$ and MGTs (unattacked) to estimate the target distribution $P$.}\\
   \midrule
    \multicolumn{2}{c}{Cosine Similarity}  & $[-1, 1] \downarrow$ & Utilize BART embedding \cite{lewis-etal-2020-bart} to compare the similarity of texts after the attack to before the attack. \\
  \multirow{2}{*}[-0.7em]{\makecell{BERTScore\\ \citep{zhang2019bertscore}}} & \textcolor{lightgray}{M2H} & \textcolor{lightgray}{$[0, 1] \leftrightarrow$} &  \textcolor{lightgray}{MGTs as the candidates $\accentset{\wedge}{x}$ and HWTs as the reference $x$. For attacked scenarios, the closer value to the unattacked scenario is favored.}\\
   & A2B & $[0, 1] \downarrow$ &  MGTs (attacked) as the candidates $\accentset{\wedge}{x}$ and MGTs (unattacked) as the reference $x$. \\
  \multirow{2}{*}[-0.5em]{\makecell{BARTScore\\\citep{yuan2021bartscore}}} & \textcolor{lightgray}{M2H} & \textcolor{lightgray}{$<0 \leftrightarrow$} &  \textcolor{lightgray}{MGTs as the source $x$ and HWTs as the target $y$. For attacked scenarios, the closer value to the unattacked scenario is favored.}\\
   & A2B & $ <0 \downarrow$ &  MGTs (attacked) as the source $x$ and MGTs (unattacked) as the target $y$. \\
 \bottomrule
\end{tabular}
}
\caption{\textbf{The metrics considered to evaluate the budget of attacks.} $\uparrow$ means a larger number represents a more significant attack on the raw texts. $\leftrightarrow$ means the value closer to the value of unattacked texts is favorable. `M2H' is `MGT to HWT,' and `A2B' is `After to Before Attack' for short. Metrics in \textcolor{lightgray}{grey} are not distinguishable enough empirically that we do not show in the paper, but are also implemented and reported in our code and data repertory.}
\label{tab:budgets}
\end{table*}

\end{document}